\documentclass[runningheads]{llncs}

\usepackage{eccv}

\usepackage{eccvabbrv}

\usepackage{graphicx}
\usepackage{booktabs}
\usepackage{tabularx}

\usepackage{microtype}
\usepackage{afterpage}
\usepackage[accsupp]{axessibility}  
\newcommand{\mycheckmark}{\ding{52}}
\newcommand{\mycrossmark}{\ding{56}}
\newcolumntype{Y}{>{\centering\arraybackslash}X}

\usepackage{hyperref}

\usepackage{orcidlink}

\usepackage{multirow}
\usepackage{pifont}
\usepackage{wrapfig}
\usepackage{sidecap}
\usepackage{xcolor} 
\usepackage{array}
\usepackage{colortbl}

\definecolor{carcolor}{HTML}{00008E}
\definecolor{ridercolor}{HTML}{FF0000}

\begin{document}

\title{DGInStyle: 
Domain-Generalizable Semantic Segmentation with Image Diffusion Models and Stylized Semantic Control} 

\titlerunning{DGInStyle}

\author{Yuru Jia\inst{1,2} \and
Lukas Hoyer\inst{1} \and
Shengyu Huang\inst{1} \and
Tianfu Wang\inst{1} \and
Luc Van Gool\inst{1,2,3} \and
Konrad Schindler\inst{1} \and
Anton Obukhov\inst{1}
}

\authorrunning{Y.~Jia et al.}

\institute{$^1$ETH Z\"{u}rich, Switzerland \quad $^2$KU Leuven, Belgium \quad $^3$INSAIT Sofia, Bulgaria}
\maketitle 

\renewcommand\twocolumn[1][]{#1}%
\begin{center}
  \includegraphics[width=\textwidth]{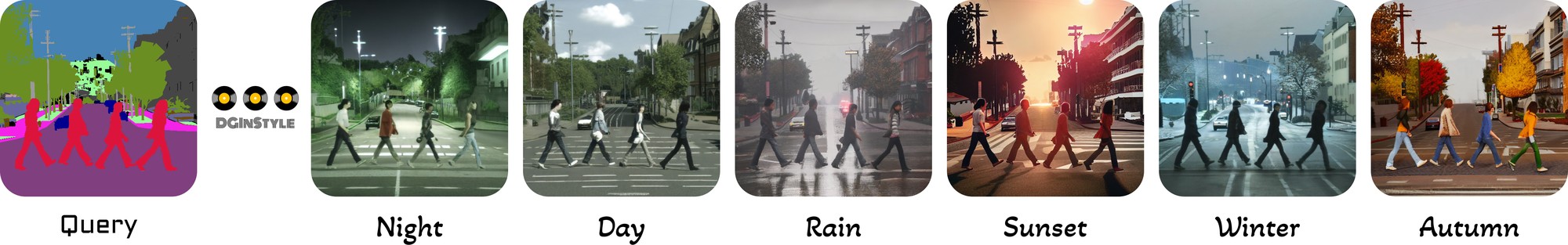}
  \captionsetup{type=figure}
  \captionof{figure}{
    \textbf{Crossing domain boundaries with DGInStyle.}
    We propose a data-centric generative pipeline for domain generalization.
    It is derived from Stable Diffusion and augmented with a novel high-precision style-preserving semantic control.
    DGInStyle combines semantic masks (\texttt{Query}) with style prompts (e.g., \texttt{Night} or \texttt{Rain}) to generate training data for semantic segmentation networks with widely varying appearance.
    It achieves state-of-the-art semantic segmentation across domains in autonomous driving.
  }
  \label{fig:teaser}
 \end{center}%

\begin{abstract}
Large, pretrained latent diffusion models (LDMs) have demonstrated an extraordinary ability to generate creative content, specialize to user data through few-shot fine-tuning, and condition their output on other modalities, such as semantic maps.
However, are they usable as large-scale data generators, e.g., to improve tasks in the perception stack, like semantic segmentation?
We investigate this question in the context of autonomous driving, and answer it with a resounding "yes".
We propose an efficient data generation pipeline termed \emph{DGInStyle}. 
First, we examine the problem of specializing a pretrained LDM to semantically-controlled generation within a narrow domain.
Second, we propose a Style Swap technique to endow the rich generative prior with the learned semantic control.
Third, we design a Multi-resolution Latent Fusion technique to overcome the bias of LDMs towards dominant objects.
Using DGInStyle, we generate a diverse dataset of street scenes, train a domain-agnostic semantic segmentation model on it, and evaluate the model on multiple popular autonomous driving datasets.
Our approach consistently increases the performance of several domain generalization methods compared to the previous state-of-the-art methods. The source code and the generated dataset are available at \href{https://dginstyle.github.io/}{dginstyle.github.io}.  
\keywords{
    Semantic Domain Generalization \and 
    Image Latent Diffusion
}
\end{abstract}

\section{Introduction}
\label{sec:intro}
The rise of generative image modeling has been a game changer for AI-assisted creativity.
Moreover, it also paves the way for improvements beyond artistic generation, particularly in computer vision.
In this paper, we investigate one such avenue and use a powerful text-to-image generative diffusion model to improve the robustness of semantic segmentation with respect to domain shifts. 

Segmenting images into semantically defined categories requires large annotated datasets of images and associated label maps, as a basis for supervised training.
Manual annotation for obtaining those label maps is time-consuming and expensive~\cite{cordts2016cityscapes,sakaridis2021acdc}, which is where image generation comes into play. Synthetic datasets are annotated by construction and therefore cheap to collect, but they invariably suffer from a \textit{domain gap}~\cite{ganin2015unsupervised}, meaning that a network trained on such data (the \textit{source domain}) will perform poorly on the real images of interest (the \textit{target domain}).
When the characteristics of the target domain are known in advance through (labeled or unlabeled) samples, the domain gap can be addressed with Domain Adaptation techniques~\cite{ganin2015unsupervised, hoyer2022daformer}.
A more challenging, arguably equally important setting is Domain Generalization (DG)~\cite{zhao2022style,hoyer2023domain,ding2023hgformer}, where a model is deployed in a new environment without the chance to first collect data and adapt. I.e., the target domain is unknown except for the high-level application context (such as ``autonomous driving'').

In the DG semantic segmentation literature, the role of the \textit{prior domain} is often overlooked. In end-to-end pipelines, that prior typically remains implicit; for instance, it could stem from pretrained backbone weights used in most segmentation DG methods (often from ImageNet~\cite{deng2009imagenet}) or loss functions that depend on feature space distances~\cite{hoyer2023domain,zhao2022style}.
Therefore, we take a closer look at the prior domain and study how we can utilize the rich prior that emerges in modern foundational models trained on internet-scale datasets~\cite{schuhmann2022laion} to improve domain generalization of semantic segmentation.

To this end, we design \mbox{\textbf{DGInStyle}}, a novel data generation pipeline with a pretrained foundational text-to-image LDM~\cite{rombach2022high} at its core, fine-tuned with data from the source domain, with conditioning on the associated dense label maps.
Such a pipeline can automatically generate images with \emph{characteristics of the prior domain} and \emph{equipped with pixel-aligned label maps} (Fig.~\ref{fig:teaser}).
Armed with such a pipeline, we approach DG differently from other methods by focusing on synthesizing data instead of model architectures or training techniques. 
The idea is that a model trained on such data will offer improved domain generalization, drawing on the prior knowledge embedded in the LDM.

This comes with two important new challenges: The LDM needs to learn how to produce images that match semantic segmentation masks. This can only be learned on the labeled source domain. However, during the process, the LDM must not overfit to the source domain style. Additionally, the generated images must exactly align with segmentation masks, even for very small instances. 
\begin{SCfigure}
    \centering
    \includegraphics[width=0.54\linewidth,height=0.305\linewidth,trim={0em 0em 0em 0em},clip]{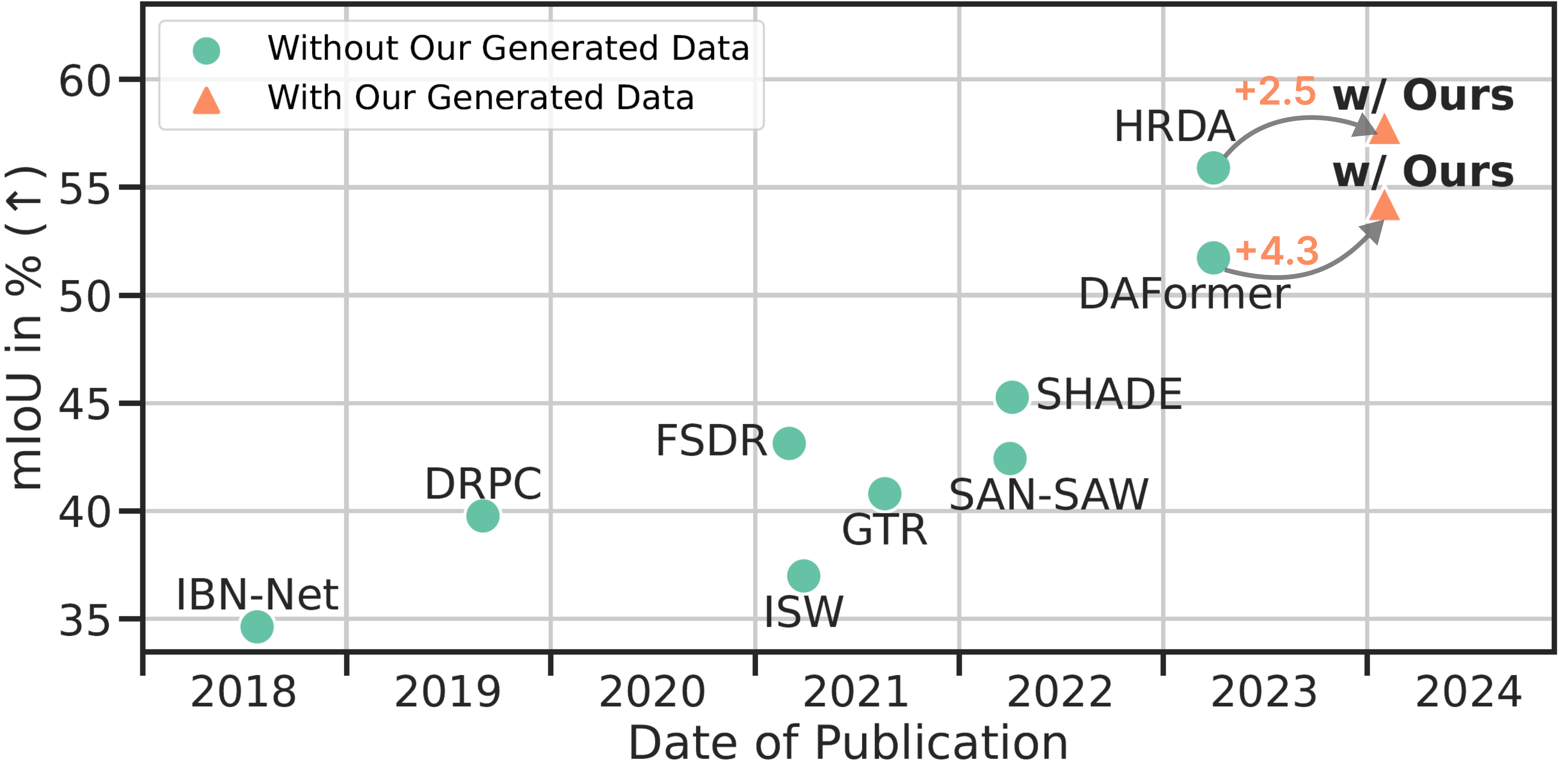}
    \caption{
        \textbf{A historical view} of domain generalization (DG) in semantic segmentation.
        The $y$-axis shows average mIoU values over three autonomous driving benchmarks: Cityscapes~\cite{cordts2016cityscapes}, BDD100K~\cite{yu2020bdd100k}, and Mapillary Vistas~\cite{neuhold2017mapillary}.
        Our data generation pipeline markedly raises the performance of high-performing  DG methods like DAFormer~\cite{hoyer2022daformer,hoyer2023domain} or HRDA~\cite{hoyer2022hrda,hoyer2023domain}.
        }%
    \label{fig:history}
\end{SCfigure}%
Therefore, several fundamental modifications are necessary to turn an off-the-shelf LDM~\cite{rombach2022high} into a data generation pipeline for domain-generalizable semantic segmentation, which would otherwise suffer from source domain style bleeding and ignoring small instances.
Our \textbf{Contributions} address these issues: \textbf{First} we propose a novel Style Swap technique inspired by modern fine-tuning and semantic style control mechanisms, to achieve the necessary level of control and diversity over the outputs. It is based on the novel finding that the semantic control and the underlying (stylized) diffusion model can be decoupled and swapped. This enables our simple yet efficient Style Swap, which allows to learn dense semantic control on the source domain while removing the undesired source domain style. 
\textbf{Second}, we present a novel Multi-Resolution Latent Fusion technique, which helps us to go beyond the limited resolution of the LDM generator. It is an essential step to achieve conditioned generation of small instances, which is crucial for learning semantic segmentation on generated data. Without it, a segmentation model trained on it would struggle with inconsistent generated images and segmentation masks.  
\textbf{Lastly}, we use the resulting generative pipeline to create a diversified dataset to train semantic segmentation networks, including methods to mitigate the impact of domain shifts. 
Due to its complementary design, DGInStyle achieves major performance improvements when combined with existing DG methods. In particular, it significantly boosts the state-of-the-art domain generalization in autonomous driving, as shown in Fig.~\ref{fig:history}.

\section{Related Work}
\label{sec:related_work}

Deep neural networks require extensive training data, which can be costly and time-consuming to acquire. 
Data access and usage scenarios are severely regulated in some domains, such as medical imaging. 
To mitigate this, there has been a growing interest in synthetic datasets.
Due to the inevitable domain gap between synthetic datasets and application scopes, domain adaptation methods focusing on a single target domain, or domain generalization, focusing on the wider task-specific domain, come to the rescue.
Creating a realistic synthetic dataset often involves physically-based simulators (e.g., renderers~\cite{roberts2021hypersim}), which also turns out expensive, and a challenge in its own right.
Thus, the recent trend of leveraging generative models for realistic data generation is winning in cost efficiency.

\vspace{1mm}
\noindent\textbf{Generative Models.}
Early advancements in deep learning techniques led to a surge in deep generative models, namely GANs and VAEs~\cite{goodfellow2014generative,kingma2022autoencoding,rezende2016variational,cai2022pix2nerf}.
While GANs exhibited training challenges such as instability and mode collapse~\cite{brock2019large}, VAEs struggled with output quality and artifacts. 

Diffusion Models (DMs)~\cite{ho2020denoising,dhariwal2021diffusion,song2021scorebased,meng2021sdedit,song2022denoising,cai2023diffdreamer} have recently demonstrated state-of-the-art image generation quality, which happened thanks to the simplified training objective, enabling training on large-scale datasets. 
These models involve a forward diffusion process that adds noise into data and a learned reverse process that recovers the original data. 
To reduce the computational requirements, latent diffusion models (LDMs)~\cite{rombach2022high} operate in the latent space, thus enabling absorption of internet-scale data~\cite{schuhmann2022laion}.
Additionally, advances in image captioning and language conditioning, such as CLIP~\cite{radford2021learning}, enabled text-guided control of the generation process. 
These advancements suggest the emergence of strong scene-understanding prior, which can be utilized for in-domain data generation.

Despite their large size, DreamBooth~\cite{ruiz2023dreambooth} demonstrated that LDMs can be efficiently fine-tuned. 
To further enhance the controllability beyond text prompts, a variety of diffusion models \cite{ho2022classifierfree,mou2023t2iadapter,zhang2023adding,huang2023composer,zhao2023unicontrolnet,goel2023pairdiffusion,ham2023modulating} integrate additional inputs of condition signals to provide more granular control.
As demonstrated in~\cite{ke2023repurposing}, LDMs can be repurposed to learn new tasks through fine-tuning and extra conditioning.
The usage of segmentation masks to guide generation has been a focal point of research, with methodologies primarily falling into condition-based \cite{zhang2023adding,mou2023t2iadapter,huang2023composer} and guidance-based categories \cite{bansal2023universal,yu2023freedom,xue2023freestyle,chen2023trainingfree}. 
When using pretrained off-the-shelf models, the limited resolution of LDMs can be an obstacle to large-scale high-resolution data generation. 
Yet, it can also be worked around, as studied in panorama generation literature~\cite{bartal2023multidiffusion}.
These techniques offer precise pixel-wise control and, subsequently, a means of generating image-label pairs for downstream tasks.

\vspace{1mm}
\noindent\textbf{Dataset Generation.}
The pioneering work DatasetGAN~\cite{zhang2021datasetgan} automatically generates labeled images by manipulating feature maps from StyleGAN and outperforms semi-supervised baselines in object-part segmentation. 
Recent techniques have utilized state-of-the-art DMs to create training data for downstream tasks such as image classification \cite{he2023synthetic,trabucco2023effective,dunlap2023diversify,sariyildiz2023fake,azizi2023synthetic,li2023synthetic}, object detection~\cite{zhang2023diffusionengine,chen2023geodiffusion,wu2022synthetic}, semantic segmentation \cite{wu2023datasetdm,kondapaneni2023textimage,peng2023diffusion,yang2023freemask,gong2023prompting,wu2023diffumask}. 

Approaches to paired image-mask dataset generation can be categorized into three groups. 
The first approach falls into the category of \textit{grounded generation} \cite{li2023guiding,kondapaneni2023textimage,xu2023openvocabulary,wu2023datasetdm,gong2023prompting}, which generates masks with the help of a separate segmentation decoder. 
This often involves a pretrained off-the-shelf network, and the domain it is trained on introduces additional biases bleeding into the overall generation process.
The second approach falls under the umbrella of \textit{image-to-image translation} techniques.
\cite{peng2023diffusion} use a DM to progressively transform images from the synthetic source domain into images resembling the target domain, guided by the source domain masks. 
The third approach uses semantic masks to guide the image generation (\textit{Semantic guidance})~\cite{yang2023freemask,zhang2023adding,mou2023t2iadapter}. 
While arguably cleaner, it also comes with caveats: the unknown distribution of masks and the degree of agreement between the generation result and the mask condition.
DGInStyle follows into the last category.
We use masks from the source domain and enforce the generation fidelity using the proposed Multi-Resolution Latent Fusion technique.

\vspace{1mm}
\noindent\textbf{Domain Generalization.}
Unsupervised Domain Adaptation (UDA)~\cite{ganin2015unsupervised} focuses on learning to perform a task in a target domain through supervised learning on labeled data from a similar source domain.
Only unannotated data from the target domain is available in this setting.
This task received much attention due to the simplicity of the formulation; several approaches~\cite{hoyer2022hrda,Saha_2023_ICCV} were proposed to efficiently bridge the domain gap. 

Domain Generalization (DG) aims to enhance the robustness of models trained on source domains and enable them to perform well on unseen domains belonging to the same task group. 
Compared to UDA, no data from the target domain is available along with the domain itself; it is defined through a union of task-specific proxy evaluation datasets.
To improve domain generalization in semantic segmentation, prior methods utilize transformations such as instance normalization~\cite{pan2018two} or whitening~\cite{choi2021robustnet} to align various source domain data with a standardized feature space.
Another line of research~\cite{zhao2022style,yue2019domain,peng2021global,zhong2022adversarial,huang2023style,kim2023texture} focuses on domain randomization, which augments the source domain with diverse styles. 
For instance, \cite{zhao2022style} selects basis styles from the source distribution, enabling the model to generate diverse samples during training. HGFormer~\cite{ding2023hgformer} improves the robustness of networks by introducing a hierarchical grouping mechanism that groups pixels to form part-level and whole-level masks for class label prediction. 
Fig.~\ref{fig:history} shows the progress in domain generalization over recent years measured on the task of autonomous driving scene segmentation.
Improvements achieved with our approach and SOTA techniques are clearly demonstrated.

\vspace{1mm}
\noindent\textbf{Diffusion Models for Domain Generalization.}
Beyond the aforementioned approaches, recent works have explored the use of diffusion models for domain generalization in semantic segmentation. 
Gong~\etal~\cite{gong2023prompting} investigate how well diffusion-pretrained representations generalize to new domains and introduce a prompt randomization strategy to enhance cross-domain performance. 
DatasetDM~\cite{wu2023datasetdm} presents a generic dataset generation model capable of producing diverse images and corresponding annotations using diffusion models. These methods implement grounded generation by training a segmentation decoder to achieve image-mask alignment.
Our approach takes a different semantic guidance route, exhibiting higher controllability and generating consistent image-label pairs that qualify as training datasets.

\section{Methods}
\label{sec:method}

Domain generalization for semantic segmentation aims to learn a model that is robust to unseen task domains using only annotated source domain data.
In this work, given the labeled source domain represented as $\mathbf{D}^\mathcal{S}=\left\{\left(x_i^\mathcal{S}, y_i^\mathcal{S}\right)\right\}_{i=1}^{N_\mathcal{S}}$, the goal is to generalize the semantic segmentation model $f_\theta$ to unseen target domains $\mathbf{D}^\mathcal{T}$ from the same task group, by utilizing the generated labeled dataset $\mathbf{D}^\mathcal{G}=\left\{\left(x_i^\mathcal{G}, y_i^\mathcal{G}\right)\right\}_{i=1}^{N_\mathcal{G}}$ in style of the prior domain $\mathcal{P}$ (hence DGInStyle), thus maximizing the overlap with the target domain.
In these notations, $x$ and $y$ stand for the images and their corresponding labels, respectively, whereas $N_\mathcal{S}$ and $N_\mathcal{G}$ are the total number of images in each dataset. 
The set $\{y_i^\mathcal{G}\}_{i=1}^{N_\mathcal{G}}$ is a subset of $\{y_i^\mathcal{S}\}_{i=1}^{N_\mathcal{S}}$ in our case, although other labels are possible.

\subsection{Label Conditioned Image Generation}
\label{sec:method_controlnet}

The success of pretrained text-to-image latent diffusion models, e.g., Stable Diffusion~\cite{rombach2022high}, provides opportunities for generating additional data to train deep neural networks. 
An LDM contains a U-Net~\cite{ronneberger2015unet} denoiser and a variational auto-encoder (VAE)~\cite{kingma2022autoencoding} to represent images in a low-resolution latent space, significantly reducing computational cost during training.
However, the generated images have no corresponding semantic segmentation mask, which is necessary for DG training.
We use existing semantic masks and conditional image generation to obtain pairs of pixel-aligned images and masks.

Specifically, we employ the recent work ControlNet~\cite{zhang2023adding} due to its efficient guidance and accessible computational requirements. 
ControlNet injects conditional inputs into the denoising process through an additional module that directly reuses the encoding layers and their weights from the base LDM. 
It connects the neural architecture via zero convolutions to enable fast fine-tuning. 
During training, we convert segmentation masks into one-hot encodings, pass them as inputs to ControlNet, and supervise it with the corresponding images from the source domain.
We also pass the unique class names extracted from the segmentation mask as a text prompt. 
Once trained, we condition the generation process on source domain masks and thus construct the new training data. 
 

\subsection{Preserving Style Prior with Style Swap}
\label{sec:method_stylization}
When training ControlNet starting from the base LDM pretrained on the prior domain, we observe that the model not only learns the mask-image alignment but also tends to overfit to the style of the domain it is fine-tuned on, as shown in \cref{fig:controlnet-style} (c).
This is undesirable as it restricts the diversity of styles in the generated images, which is critical to domain generalization. 
\noindent 
\begin{figure}[b]
\begin{minipage}[t]{0.48\textwidth}
    \centering
    \scriptsize
    \includegraphics[width=\linewidth,trim={0 0 0 0},clip]{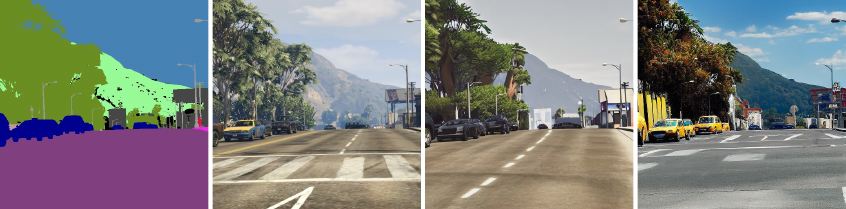}
    \begin{tabularx}{\linewidth}{*{4}{Y}}
    (a) Source Mask & (b) Source Image & (c) Gen. w/o Swap & (d) Gen. w/ Swap\\
    \end{tabularx}
    \caption{
    \textbf{
    ControlNet learns the source domain style.
    }
    This effect hinders varied data generation for domain generalization. Our Style Swap mitigates the effect and preserves the style prior.}
    \label{fig:controlnet-style}
\end{minipage}
\hfill
\begin{minipage}[t]{0.48\textwidth}
    \centering
    \scriptsize
    \includegraphics[width=\linewidth,trim={0 0 0 0},clip]{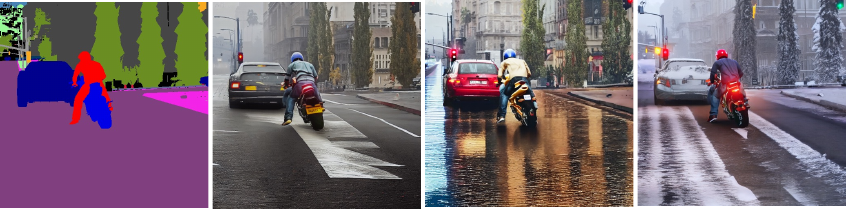}
    \begin{tabularx}{\linewidth}{*{4}{Y}}    
    (a) Car, Rider... & (b) Foggy & (c) Rainy & (d) Snowy\\
    \end{tabularx}
    \caption{
    \textbf{Style variations.}
    DGInStyle can generate images under various scene conditions through style prompting, while maintaining consistent dense semantic control from (a).
    }
    \label{fig:styles}
\end{minipage}
\end{figure}
 
\begin{figure*}[t]
    \centering
    \includegraphics[width=\linewidth,trim={0 0 0 0},clip]{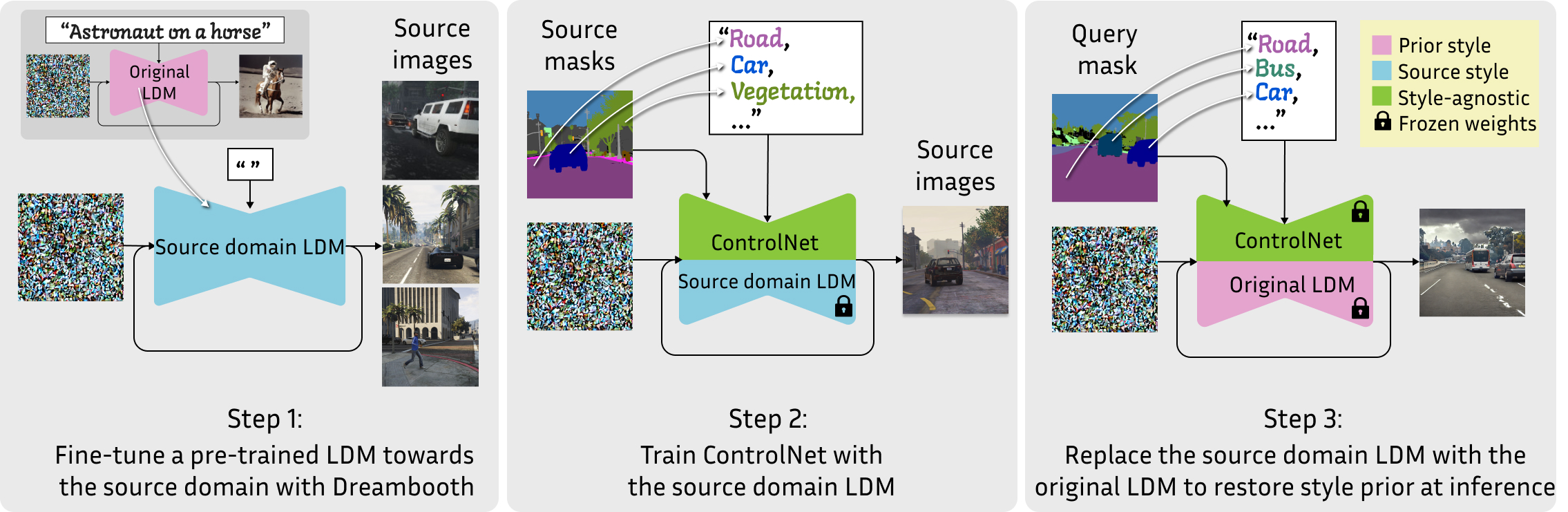}
    \caption{
        \textbf{Overview of our proposed Style Swap technique.}
        ControlNet learns segmentation-conditioned image generation on the source domain. To avoid ControlNet steering the generated style, it is trained on top of a source domain fine-tuned LDM.
        Later, this source domain LDM can be replaced with the original LDM to restore the rich style prior. 
        As discussed in Sec.~\ref{sec:experiments}, this technique leads to state-of-the-art results in domain generalization for semantic segmentation.
    }
\label{fig:style_swap}
\end{figure*}
To mitigate this issue, we develop a Style Swap technique to remove the domain-specific style and achieve diverse stylization by retrieving the prior knowledge baked in the pretrained LDMs in three steps, shown in Fig.~\ref{fig:style_swap}.

DreamBooth~\cite{ruiz2023dreambooth} was originally proposed as an efficient protocol for few-shot fine-tuning of a pretrained LDM to learn a new concept, represented in the training images and unavailable in the prior domain.
We employ its reconstruction loss as an efficient means for fine-tuning the LDM towards whole domains.

As a first step of our Style Swap technique, we fine-tune the base LDM's U-Net $\mathbf{U}^{\mathcal{P}}$ encapsulating the prior domain $\mathcal{P}$ with the Dreambooth protocol~\cite{ruiz2023dreambooth} using all images in the source domain $\mathcal{S}$.
The resulting U-Net is denoted as $\mathbf{U}^{\mathcal{S}}$. 
Second, we use $\mathbf{U}^{\mathcal{S}}$ as the base model instead of $\mathbf{U}^{\mathcal{P}}$ to initialize ControlNet.
The idea is to allow $\mathbf{U}^{\mathcal{S}}$ to absorb the domain style and let the ControlNet focus primarily on the task-specific yet style-agnostic layout control, thereby reducing the domain style bleeding into its weights.
Finally, in the third step, we perform inference with the trained ControlNet, except that we switch the base LDM generator to $\mathbf{U}^{\mathcal{P}}$ while keeping the ControlNet trained for $\mathbf{U^{\mathcal{S}}}$. 
This enables us to apply semantic control learned from the source domain to the original LDM.
The overall procedure endows the original LDM with task-specific semantic control, allowing us to generate diverse images adhering to the semantic segmentation masks.
This result is shown in Fig.~\ref{fig:controlnet-style} (d).


\subsection{Style Prompting}
\label{sec:method_prompts}

Text prompting is a powerful technique for style mining.
To better guide ControlNet generation, we concatenate unique class names present in the semantic mask into a list and pass it to the text encoder. 
We further enrich the diversity of the generated data by fusing randomized task-specific qualifiers into the text conditioning. These can be obtained from the task definition with a query to a domain expert or an LLM, and sometimes are known in advance, e.g., from the source data simulator, such as GTA~\cite{richter2016playing}. 
For the autonomous driving segmentation, we use a range of adverse weather conditions (e.g., foggy, snowy, rainy, overcast, and night scenarios). 
An example text prompt can be: \textit{A city street scene photo with car, road, sky, rider, bicycle, vegetation, building, in foggy weather}.  
This approach, especially when integrated with the Style Swap technique, allows producing images with 
semantic control and varied natural styles from the prior domain $\mathcal{P}$, as shown in~\cref{fig:styles}.

\subsection{Multi-Resolution Latent Fusion}
\label{sec:method_multidiff}

\begin{figure*}[t!]
    \centering
    \includegraphics[width=\linewidth,trim={0 0 0 0},clip]{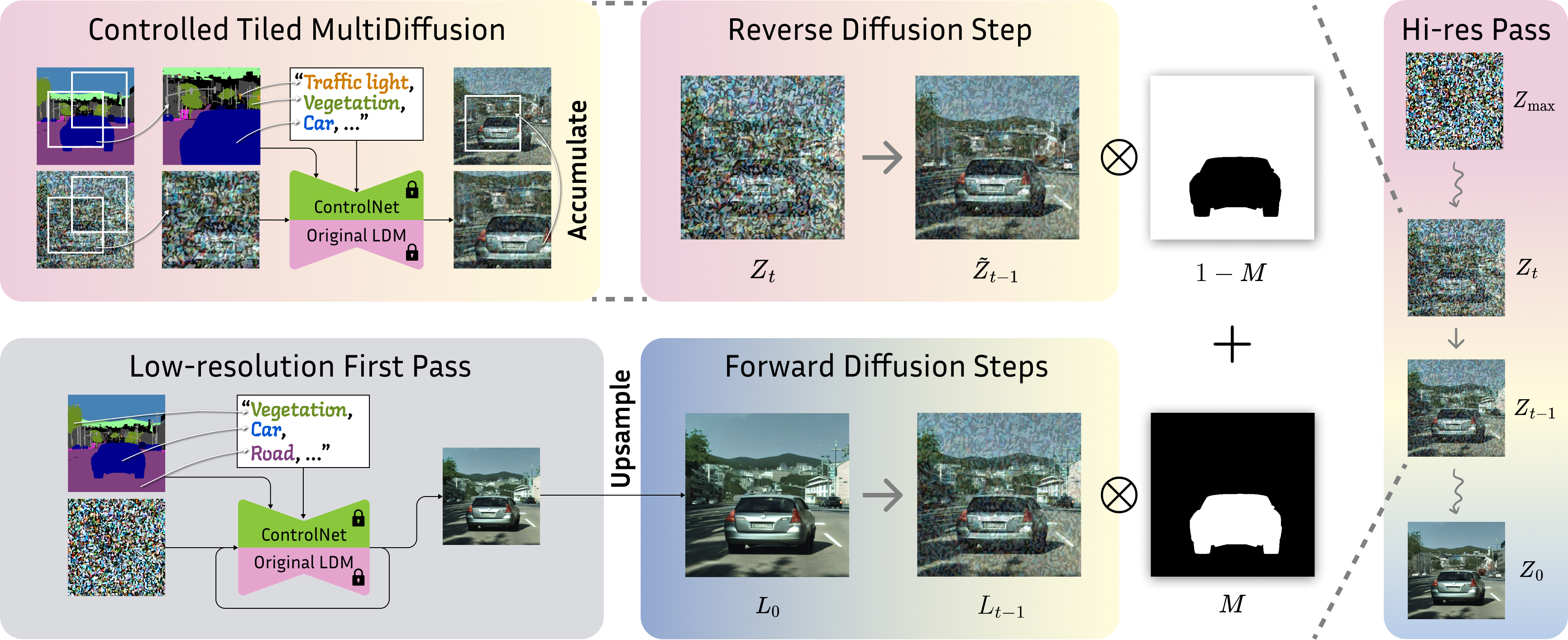}
    \caption{
    \textbf{MRLF module.} 
    We generate a first-pass image $I$ using low-resolution conditioning. 
    In the subsequent high-resolution pass, we partition the canvas into overlapping tiles at each generation step, concurrently apply denoising to each with its respective conditioning, and fuse them with a tile diffusion technique. 
    Finally, we preserve the quality of large objects in the mask $\mathrm{M}$ with inpainting conditioned on the first pass image. The color gradient \ \raisebox{-0.08em}{\includegraphics[width=1.0cm,height=0.25cm]{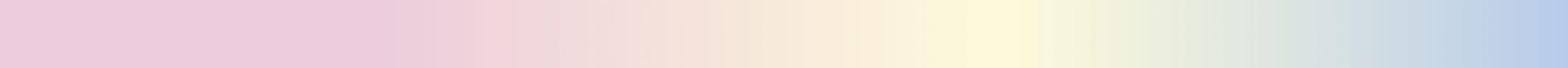}}\ \  represents the path from noise to clean data.
    }
    \label{fig:multidiff_method}
\end{figure*}

While ControlNet effectively integrates condition masks into the generation process, it struggles with generating small objects due to the low-resolution latent space.
We propose a two-stage Multi-Resolution Latent Fusion pipeline to improve the adherence to semantic masks in the generated dataset. 
During the first low-resolution pass (Fig.~\ref{fig:multidiff_method}, bottom-left), we perform a regular ControlNet generation at the original LDM resolution.
This generation serves as a reference for the second, high-resolution generation pass.
Therein, we keep the large segments generated initially and refine everything else.
To overcome the problem of low resolution of the latent space, we perform the second pass in the upsampled latent space, followed by downsizing the generated image to the original size.
Such a two-stage pipeline makes use of two other techniques, specifically, the Controlled Tiled MultiDiffusion (Fig.~\ref{fig:multidiff_method}, top-left), and the Latent Inpainting Diffusion, seen on the right side of the figure.

\vspace{1mm}
\noindent\textbf{Controlled Tiled MultiDiffusion.}
We choose an upscaling factor $s$ and initialize the high-resolution latent code $Z \in \mathbb{R}^{sw \times sw \times d}$ with Gaussian noise, where $w \times h \times d$ is the resolution of the denoiser U-Net input.
The condition mask $y$ is upsampled to $Y$ by the same factor $s$ using nearest-neighbor interpolation.
Next, the latent canvas $Z$ is divided into a grid of regularly spaced overlapping tiles of size $w \times h \times d$ for subsequent diffusion.

To perform a single diffusion update step $t$ over the whole canvas, we crop tuples of intermediate latent codes and their corresponding spatially-aligned conditions $(Z_{i,t}, Y_i), i=1,\ldots,n$ and perform the standard controlled denoising step update with ControlNet discussed previously for each of them independently. 
As with the low-resolution pass, we condition each crop denoising step on the relevant set of semantic classes and the style prompt. 
Next, we paste the updated latent codes back into the canvas.
The overlapping tiles are fused by averaging overlapping areas following MultiDiffusion~\cite{bartal2023multidiffusion}.

Such a controlled generation in the upsampled space overcomes the low-resolution bias of the pretrained LDM and results in higher-quality small objects.
Nevertheless, this procedure alone leads to a noticeable degradation of large objects due to the reduced field of view of the denoiser. 
We address this shortcoming by taking large objects from the first low-resolution pass and fusing them into the high-resolution pass using the Latent Inpainting Diffusion technique.

\vspace{1mm}
\noindent\textbf{Latent Inpainting Diffusion.}
To detect large areas to keep from the first pass, we perform connected component analysis of the original segmentation masks. 
Large components with a relative area over a certain threshold contribute to the binary mask $M \in \mathbb{R}^{sh\times sw}$, compatible in dimensions with the latent canvas $Z$.
After extracting these large component regions, we generate the high-resolution image using a modified diffusion pipeline, similar to~\cite{lugmayr2022repaint, wang2023breathing}.
First, we perform Controlled Tiled MultiDiffusion at each step to deal with the latent canvas.
Second, we compose the final latents on step $t-1$ from the denoised latents $\tilde{Z}_{t-1}$ (from $Z_{t}$) and the low-resolution outcome.
Specifically, we upsample the low-resolution image, encode it into the enlarged latent space using VAE to get $L_0$, and apply the forward diffusion process to obtain the latent code at step $L_{t-1}$.
The resulting latent canvases of compatible dimensions are blended using the mask $M$: $Z_{t - 1}= \left(1-M\right) \otimes \Tilde{Z}_{t-1} + M \otimes{L_{t - 1}}$.

As a result, our multi-resolution latent fusion scheme overcomes the resolution-specific limitations of the LDM.
It unlocks controlled arbitrary-resolution generation through processing tiles. At the same time, it preserves trusted regions with the latent inpainting diffusion scheme.

\subsection{Rare Class Generation}
\label{sec:method_rcg}
Perception models trained on imbalanced datasets tend to be biased towards common classes and perform poorly on rare classes. 
We address this challenge by considering class distribution at both the ControlNet training and final dataset generation phases.

Specifically, for each class $c$ with frequency $f_c$ in the source domain, its sampling probability is 
$P(c)=e^{\left(1-f_c\right) / T} / \sum_{c^{\prime}=1}^C e^{\left(1-f_{c^{\prime}}\right) / T}$,
where $C$ is the total number of classes, and $T$ controls the smoothness of the class distribution.
During the training phase of ControlNet, we prioritize and sample more frequently those image-mask pairs featuring rare classes.
This helps ControlNet recognize and handle these challenging classes. During the dataset generation phase, we increase the frequency of choosing semantic masks containing rare classes to boost the proportion of rare classes in the generated dataset.

\section{Experiments}
\label{sec:experiments}

\subsection{Datasets}
Following the common practice in domain generalization literature~\cite{hoyer2022daformer,hoyer2023domain}, we use \textbf{GTA}~\cite{richter2016playing} with a total of 24966 images as the synthetic source dataset. 
To evaluate our method's domain generalization capability, we employ five real-world datasets within the context of autonomous driving.
\textbf{Cityscapes (CS)}~\cite{cordts2016cityscapes} is an urban street scene dataset collected in several cities in and around Germany. 
\textbf{BDD100K (BDD)}~\cite{yu2020bdd100k} contains images of urban scenes captured at different locations in the United States. 
\textbf{Mapillary Vistas (MV)}~\cite{neuhold2017mapillary} includes the world-wide street scenes and is diverse in terms of weather conditions, seasons, and daytime variations.  
Specifically for adverse conditions, we also utilize \textbf{ACDC}~\cite{sakaridis2021acdc} and \textbf{Dark Zurich (DZ)}~\cite{dz19}, both of which contain images captured under challenging weather conditions and during nighttime.

\subsection{Implementation Details}

Our model is based on Stable Diffusion 1.5~\cite{rombach2022high} and requires a single consumer-grade GPU for training.
We first conduct DreamBooth~\cite{ruiz2023dreambooth} fine-tuning using GTA images to obtain $\mathbf{U}^{\mathcal{S}}$. 
The images are randomly resized and cropped to a resolution of $512{\times}512$. 
The fine-tuning takes 10k iterations with a constant learning rate of $2{\times}10^{-6}$. 

The ControlNet~\cite{zhang2023adding} training is initialized with the source style $\mathbf{U}^{\mathcal{S}}$.
For input conditions, we use one-hot encoded GTA segmentation masks and crop them to the size of $512{\times}512$. 
These crops are guided by input text containing semantic classes in each crop.
During ControlNet inference, we perform the Style Swap as discussed in ~\cref{{sec:method_stylization}} and integrate the multi-resolution latent fusion module with $s{=}2$.
Our tiling strategy uses a 16-pixel stride between neighbor crops.
We use $T{=}0.01$ for rare class sampling probability. 
The constructed dataset comprises an equal mix of images with basic text inputs and those with randomized adverse weather prompts.
Extra examples are shown in the supplement.

To assess the efficacy of our DGInStyle, we train a semantic segmentation model on a combination of the GTA source dataset and our generated dataset. 
Specifically, we generate $N_\mathcal{G}=6000$ images and select $N_\mathcal{S}=6000$ images based on the rare class criteria.
The training is performed under the aligned domain generalization framework as detailed in~\cite{hoyer2023domain}.

\subsection{Comparison with State-of-the-Art DG}
\begin{table}[t!]
\centering
\captionof{table}{
\textbf{DG with GTA source domain and ResNet-101/MiT-B5 backbone.}
Comparison of Domain Generalization (DG) methods for semantic segmentation in autonomous driving scenes w/ and w/o integrating our generated dataset (mIoU~$\uparrow$ in \%) with GTA as source domain, using either ResNet-101 or MiT-B5 as the backbone. 
As seen, leveraging our proposed data generation pipeline, which exploits rich generative priors and semantic conditioning, provides a substantial boost in performance across various configurations.
}
\label{tab:main-mitb5}
\renewcommand{\arraystretch}{1.15}
\setlength{\tabcolsep}{3pt}
\resizebox{\columnwidth}{!}{
\begin{tabular}{@{}l c cccc c cccc@{}}

\toprule
\addlinespace[5pt]

DG Method &
DGInStyle &
CS~\cite{cordts2016cityscapes} &
BDD~\cite{yu2020bdd100k} &
MV~\cite{neuhold2017mapillary} &
Avg3 & &
ACDC~\cite{sakaridis2021acdc} &
DZ~\cite{dz19}
& Avg5 & $\Delta$Avg5\\ 

\addlinespace
\midrule

\multicolumn{11}{c}{\textbf{ResNet-101~\cite{he2016deep}}} \\
\midrule
\multirow{2}{*}{IBN-Net~\cite{pan2018two}}
& \ding{56}   & 37.37 & 34.21 & 36.81 & 36.13 & & 25.85 & 6.12 & 28.07 &\multirow{2}{*}{\textcolor{ForestGreen}{$\uparrow$ \textbf{5.1}}} \\
& \ding{52}   & \textbf{40.80} & \textbf{38.98} & \textbf{43.20} & \textbf{40.99} & & \textbf{31.68}  & \textbf{11.19}  &  \textbf{33.17} &  \\
\midrule

\multirow{2}{*}{RobustNet~\cite{choi2021robustnet}}
& \ding{56}   & 37.20 & 33.36 & 35.57 & 35.38 & & 24.80 & 5.49 & 27.28 &\multirow{2}{*}{\textcolor{ForestGreen}{$\uparrow$ \textbf{6.8}}}\\
& \ding{52}  & \textbf{41.03} & \textbf{39.62} & \textbf{44.85} & \textbf{41.83} & & \textbf{32.30}  & \textbf{12.73}  &  \textbf{34.11}  &  \\
\midrule
DRPC~\cite{yue2019domain} 
& \ding{56}   & 42.53 & 38.72 & 38.05 & 39.77 & & -- & -- & --  &  \\
FSDR~\cite{huang2021fsdr} 
& \ding{56}   & 44.80 & 41.20 & 43.40 & 43.13 & & 24.77 & \textbf{9.66}& 32.77  &  \\
GTR~\cite{peng2021global} 
& \ding{56}   & 43.70 & 39.60 & 39.10 & 40.80 & & -- & -- & --  &  \\
SAN-SAW~\cite{peng2022semantic} 
& \ding{56}   & 45.33 & 41.18 & 40.77 & 42.23 & & -- & -- & -- &  \\
AdvStyle~\cite{zhong2022adversarial} 
& \ding{56}   & 44.51 & 39.27 & 43.48 & 42.42 & & -- & -- & --  &  \\
SHADE~\cite{zhao2022style} & \ding{56}   & \textbf{46.66} & \textbf{43.66} & \textbf{45.50} & \textbf{45.27} & & \textbf{29.06} & 8.01 & \textbf{34.58}  &  \\
\midrule
\multirow{2}{*}{HRDA~\cite{hoyer2022hrda,hoyer2023domain}}
& \ding{56}   & 39.63 & 38.69 &  42.21 & 40.18 & & 26.08 & 7.80 & 30.88 &\multirow{2}{*}{\textcolor{ForestGreen}{$\uparrow$ \textbf{7.2}}} \\
 & \ding{52}   & \textbf{46.89} & \textbf{42.81} & \textbf{50.19} & \textbf{46.63} & & \textbf{34.19} & \textbf{16.16} & \textbf{38.05} \\
\midrule

\multicolumn{11}{c}{\textbf{MiT-B5~\cite{xie2021segformer}}} \\
\midrule

\multirow{2}{*}{Color-Aug} & \ding{56} & 46.64 & 45.45 & 49.04& 47.04  & & 36.10 & 16.37 & 38.72 &\multirow{2}{*}{\textcolor{ForestGreen}{$\uparrow$ \textbf{3.3}}} \\
& \ding{52} &\textbf{50.76} & \textbf{47.21} & \textbf{52.33} & \textbf{50.10} & & \textbf{38.92}  &\textbf{20.94} & \textbf{42.03}  \\

\midrule

\multirow{2}{*}{DAFormer~\cite{hoyer2022daformer,hoyer2023domain}} 
& \ding{56} & 52.65 & 47.89 & 54.66& 51.73 &  &38.25 & 17.45  & 42.18 &\multirow{2}{*}{\textcolor{ForestGreen}{$\uparrow$ \textbf{4.3}}} \\
& \ding{52} &\textbf{55.31} & \textbf{50.82} & \textbf{56.62} & \textbf{54.25} & & \textbf{44.04} & \textbf{25.58} & \textbf{46.47}\\

\midrule

\multirow{2}{*}{HRDA~\cite{hoyer2022hrda,hoyer2023domain}}
& \ding{56} & 57.41 & 49.11 & 61.16 &55.90 &  & 44.04 & 20.97&46.54 &\multirow{2}{*}{\textcolor{ForestGreen}{$\uparrow$ \textbf{2.5}}}  \\
& \ding{52} & \textbf{58.63} & \textbf{52.25} & \textbf{62.47}  &\textbf{57.78}  &&\textbf{46.07} & \textbf{25.53} & \textbf{48.99} \\ 
\bottomrule

\end{tabular}
}
\end{table}

\begin{figure}[t]
    \centering
    \includegraphics[width=\linewidth,trim={0em 0.5em 0em 0em},clip]{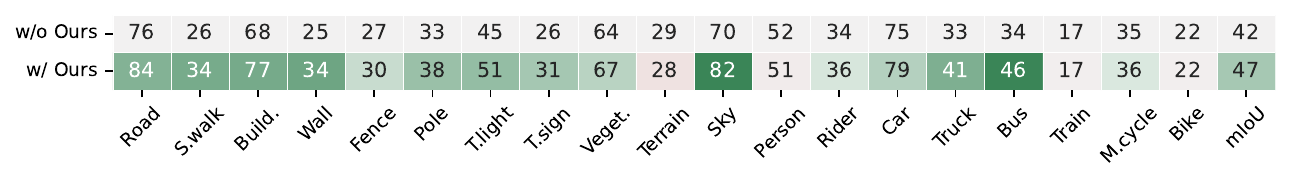}
    \caption{
    \textbf{Class-wise IoU} averaged over the five datasets using DAFormer with and without our dataset integration. The color visualizes the difference to the first row.}
    
    \label{fig:heatmap}
\end{figure}

In~\cref{tab:main-mitb5}, we benchmark several DG methods trained using either the GTA dataset alone or augmented with our DGInStyle and subsequently evaluated across five real-world datasets to measure their generalization from GTA to other domains. Specifically, we integrate DGInStyle into IBN-Net~\cite{pan2018two}, RobustNet~\cite{choi2021robustnet}, Color-Aug (random brightness, contrast, saturation, and hue), DAFormer~\cite{hoyer2022daformer,hoyer2023domain}, and HRDA~\cite{hoyer2022hrda,hoyer2023domain} covering CNN-based ResNet-101~\cite{he2016deep} and Transformer-based MiT-B5~\cite{xie2021segformer} network architectures.

The results in \cref{tab:main-mitb5} indicate that DGInStyle significantly enhances the DG performance across various DG methods and network architectures. The improvements range from +2.5 mIoU up to +7.2 mIoU on the average over 5 datasets. In particular, DGInStyle improves the overall state-of-the-art performance by a significant gain of +2.5 mIoU. These results confirms the efficacy of our method in generating diverse, style-varied image-label pairs for semantic segmentation, thereby significantly contributing to robust domain generalization across different network architectures and training strategies.

\begin{figure}[tb]
    \centering
    \begin{tabularx}{\linewidth}{*{4}{Y}}
    Image & w/o Ours & w/ Ours & Ground Truth \\
    \end{tabularx}
    \includegraphics[width=0.975\linewidth,trim={0 0 0 0},clip]{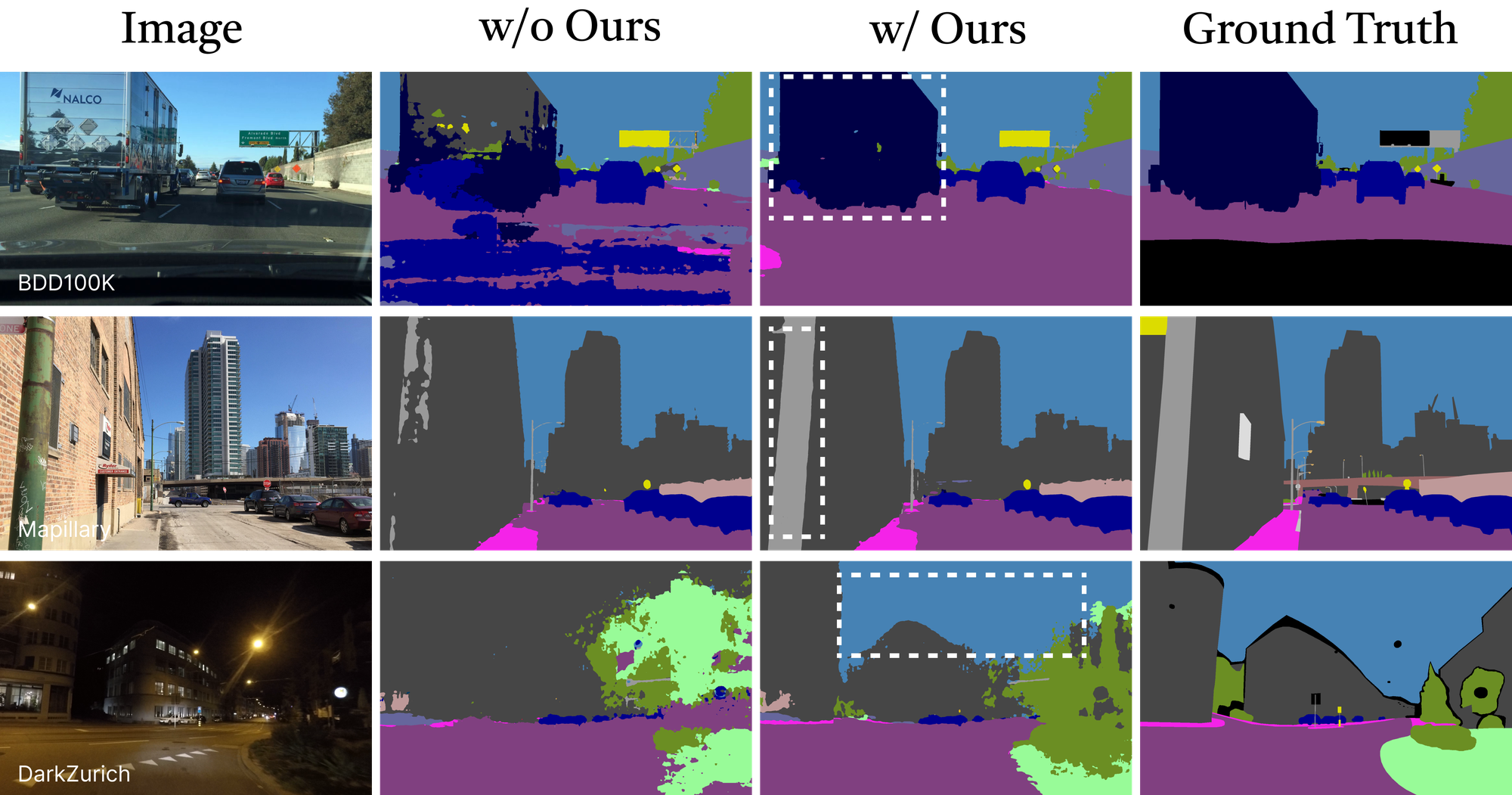}
    \caption{\textbf{Qualitative comparison} of segmentation results predicted by HRDA trained on GTA and trained on our DGInStyle. }
    \label{fig:pred-examples}
\end{figure}

We gauge the impact of our generated dataset on class-wise IoU scores using DAFormer, as shown in~\cref{fig:heatmap}. 
The heatmap affirms the capability of our data generation process across a wide range of classes. 
Notably, there is a strong improvement in classes such as \textit{pole}, \textit{traffic light}, and \textit{traffic sign}, highlighting the effectiveness of our conditioning approach, which specifically targets these small classes. 
Additionally, we observe a significant improvement in the \textit{sky} class, especially in evaluations with the DarkZurich dataset. 
This suggests that our DGInStyle is adept at bridging major domain gaps, such as transitioning to night scenes, as further exemplified in~\cref{fig:pred-examples}.

To broaden the scope of our evaluation, we set an experiment with Cityscapes \cite{cordts2016cityscapes} as a source domain, generalizing to other real-world domains in \cref{tab:cs-source-dg}. 
As a real-world dataset, Cityscapes has a smaller domain gap to the other real-world target datasets than the synthetic GTA dataset.
When using Cityscapes as a source, the baseline performance without DGInStyle is therefore naturally higher, which reduces the potential for improvement.
Yet, even in this more saturated setting, DGInStyle achieves significant average improvements.
These findings affirm the versatility and robustness of our method.
\begin{table}[t!]
\centering
\captionof{table}{
\textbf{DG with Cityscapes source domain and MiT-B5~\cite{xie2021segformer} backbone.}
Cityscapes to other datasets domain generalization w/ and w/o integrating our generated dataset (mIoU~$\uparrow$ in \%).
}
\label{tab:cs-source-dg}
\setlength{\tabcolsep}{4pt}
\resizebox{1.0\columnwidth}{!}{
\begin{tabular}{@{}l c cccc cc c @{}}

\toprule
\addlinespace[5pt]

  DG Method &
  DGInStyle &
  BDD~\cite{yu2020bdd100k} &
  MV~\cite{neuhold2017mapillary} &
  ACDC~\cite{sakaridis2021acdc} &
  DZ~\cite{dz19} & 
  Average & $\Delta$Average \\

\addlinespace
\midrule

\multirow{2}{*}{Color-Aug}& 
\ding{56}      & 53.33 & \textbf{60.06} & 52.38 & 23.00 & 47.19 &\multirow{2}{*}{\textcolor{ForestGreen}{$\uparrow$ \textbf{2.1}}}\\
 & \ding{52}   & \textbf{55.18} & 59.95 &\textbf{55.19} & \textbf{26.83} & \textbf{49.29} \\
\midrule

\multirow{2}{*}{DAFormer~\cite{hoyer2022daformer,hoyer2023domain}}  
& \ding{56} & 54.19 & 61.67  & 55.15  & 28.28 & 49.82 &\multirow{2}{*}{\textcolor{ForestGreen}{$\uparrow$ \textbf{1.5}}} \\
& \ding{52} &\textbf{56.26}  & \textbf{62.67}  & \textbf{57.74}  & \textbf{28.55} & \textbf{51.31}   \\
\midrule

\multirow{2}{*}{HRDA~\cite{hoyer2022hrda,hoyer2023domain}} 
& \ding{56} & 58.49 & \textbf{68.32}  & 59.70  & 31.07  & 54.40 &\multirow{2}{*}{\textcolor{ForestGreen}{$\uparrow$ \textbf{0.7}}}\\
& \ding{52}  & \textbf{58.84} & 67.99  & \textbf{61.00}   & \textbf{32.60} & \textbf{55.11}    \\

\bottomrule
\end{tabular}
}
\end{table}

\vspace{1mm}
\noindent\textbf{Qualitative Analysis.} \cref{fig:pred-examples} provides visual examples of semantic segmentation results obtained from HRDA trained with or without the use of DGInStyle. 
It shows that our generated dataset improves the predictions, even for difficult classes (e.g., \textit{truck} and \textit{pole}) and lighting conditions (e.g., \textit{day} and \textit{night}).
\subsection{Ablation Studies}

\begin{table}[b!]

\begin{minipage}{0.48\linewidth}
\centering
\renewcommand{\arraystretch}{1.3}
\setlength{\tabcolsep}{4pt}
\resizebox{\columnwidth}{!}{
\begin{tabular}{cccc c cc}
\toprule
\multicolumn{4}{c}{Modules} && \multicolumn{2}{c}{mIoU$\uparrow$} \\
MRLF    & Swap   & Prompts & RCG   && Avg3 & Avg5 \\ \hline
\ding{56}& \ding{56}&\ding{56}&\ding{56}&& 51.46 & 43.31 \\
\ding{52}& \ding{56}&\ding{56}&\ding{56}&& 52.84 & 44.27 \\
\ding{52}& \ding{52}&\ding{56}&\ding{56}&& 53.85 & 45.84 \\
\ding{52}& \ding{52}&\ding{52}&\ding{56}&& 53.95 & 46.16 \\ 
\ding{52}& \ding{52}&\ding{52}&\ding{52}&& \textbf{54.25} & \textbf{46.47}\\ 

\midrule

\mycrossmark & \mycheckmark & \mycheckmark & \mycheckmark && 53.07 & 45.19 \\
\mycheckmark & \mycrossmark & \mycheckmark & \mycheckmark && 51.50 & 43.12 \\
\mycheckmark & \mycheckmark & \mycrossmark & \mycheckmark && 53.85 & 44.67 \\
\mycheckmark & \mycheckmark & \mycheckmark & \mycrossmark && 53.95 & 46.16 \\
\mycheckmark & \mycheckmark & \mycheckmark & \mycheckmark && \textbf{54.25} & \textbf{46.47} \\

\bottomrule
\end{tabular}
}
\vspace{3mm}
\captionof{table}{
\textbf{Ablation studies} on different components for our data generation pipeline. All models use DAFormer~\cite{hoyer2022daformer} and are trained with GTA and our generated dataset. MRLF: our multi-resolution latent fusion module; RCG: using rare class sampling in the ControlNet training and dataset generation phases.}
\label{tab:ablation}
\end{minipage}
\hfill
\begin{minipage}{0.48\linewidth}
    \centering
    \scriptsize
    \includegraphics[width=\linewidth,trim={0 0 0 0},clip]{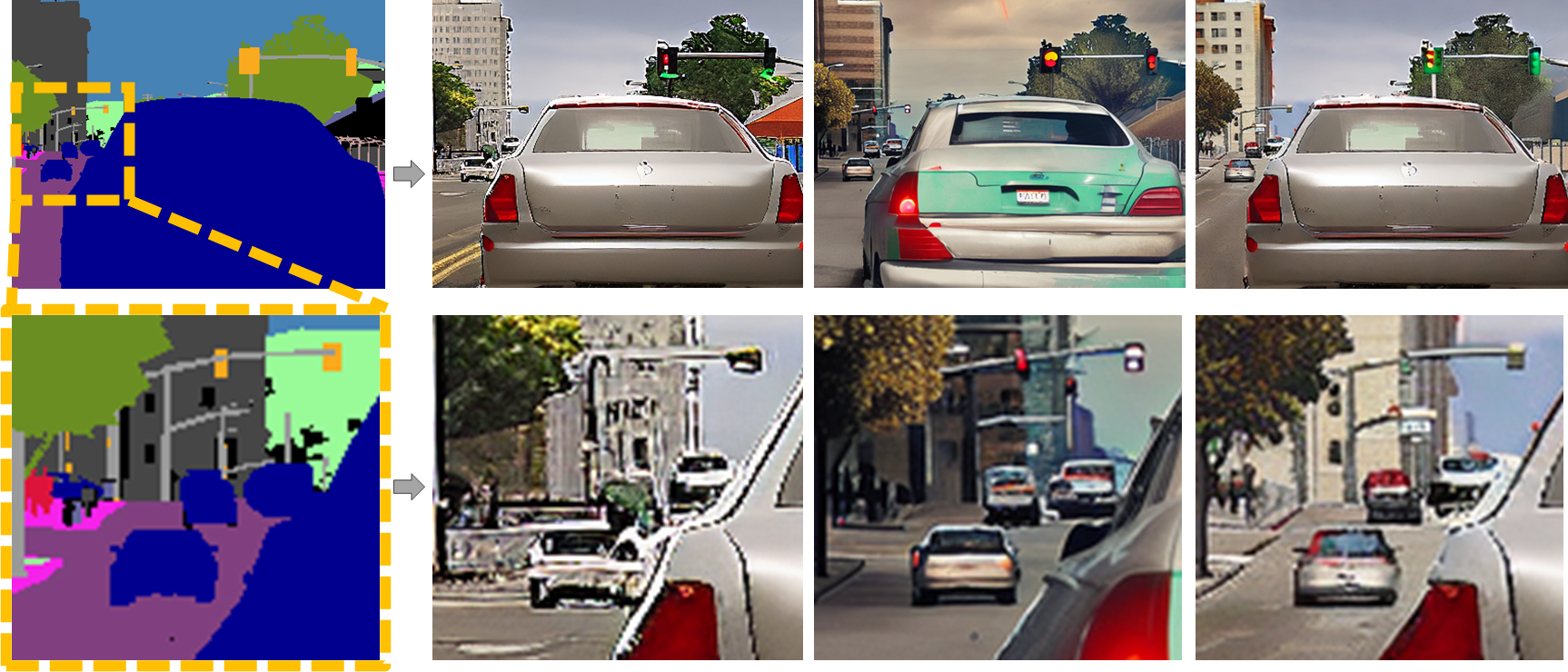}
    \begin{tabularx}{\linewidth}{*{4}{Y}}
    Semantic Mask & (a) w/o MRLF & (b) w/ CTMD & (c) w/ LID + CTMD\\
    \end{tabularx}
    \vspace{3mm}
    \captionof{figure}{\textbf{Qualitative examples of MRLF.}
    (a) When zooming in on the mask crop, which contains small objects such as cars and traffic poles, the initial generation fails to create recognizable content for these instances. (b) This is addressed by conducting Controlled Tiled MultiDiffusion, which enhances the generation quality of fine details. However, it can lead to artifacts of large objects. (c) When adding Latent Inpainting Diffusion, the generated image not only improves the local details but also reduces artifacts in large objects. 
    }
    \label{fig:multidiff-res}
\end{minipage}
\end{table}
We conduct ablation studies to evaluate the contribution of each component of our method. 
The results are shown in~\cref{tab:ablation}. 
All models are based on the DAFormer~\cite{hoyer2022daformer} framework but trained on datasets generated under varying conditions.
We observe that incorporating Multi-Resolution Latent Fusion (MRLF) enhances the generation of small objects in our dataset, boosting the segmentation model's performance by +0.96 on average of the five datasets. 
As a vital part of the style diversification module, the Style Swap technique significantly improves model performance by another +1.57, demonstrating the effectiveness of utilizing the prior domain to generate diverse samples. 
The Style Prompts module further elevates the model performance by +0.32, especially in adverse weather scenarios~\cite{sakaridis2021acdc,dz19}. 
Combined with Rare Class Generation (RCG), which adds another +0.31, our complete data generation pipeline achieves an average mIoU of 46.47\% over the five real-world datasets.

We additionally present the ablations by excluding each component during the dataset generation to evaluate their role in the combined framework.
\cref{tab:ablation} shows that removing the Style Swap component significantly degrades performance, underscoring its effectiveness in leveraging prior knowledge to diversify the generated data. 
Similarly, removing other components also leads to a decline in the model's performance, which reveals that each component adds value to our data generation pipeline.

\begin{table}[tb]
\centering
\caption{
\textbf{MRLF Ablation.}
Ablation studies on multi-resolution components with Controlled Tiled MultiDiffusion (CTMD) and Latent Inpainting Diffusion (LID). Numbers are reported in mIoU (higher is better).}
\label{tab:multi-resolution}

\scriptsize
\setlength{\tabcolsep}{16pt}
\begin{tabular}{cc c cc}
\toprule
CTMD   & LID      && Avg3      & Avg5    \\ \midrule
\ding{56}& \ding{56} && 53.07 & 45.19  \\
\ding{52}& \ding{56} && 54.05 & 45.60  \\
\ding{52}& \ding{52} && \textbf{54.25} & \textbf{46.47}  \\ 
\bottomrule
\end{tabular}

\end{table}

To gain further insights on MRLF, we ablate its two passes while incorporating all other components during dataset generation. 
As shown in~\cref{tab:multi-resolution}, both the Controlled Tiled MultiDiffusion (CTMD) and the Latent Inpainting Diffusion (LID) contribute to the overall performance of our method. This is also exemplified in~\cref{fig:multidiff-res}, where it becomes evident that the MRLF module not only refines local details but also minimizes artifacts in larger objects.

\section{Conclusion}
We have explored the potential of generative data augmentation using pretrained LDMs in the challenging context of domain generalization for semantic segmentation. 
We propose DGInStyle, a novel and efficient data generation pipeline that crafts diverse task-specific images by sampling the rich prior of a pretrained latent diffusion model, while ensuring precise adherence of the generation to semantic layout condition. 
DGInStyle has demonstrated its capability to enhance the generalizability of semantic segmentation models through extensive experiments across various domains. 
It consistently improves the performance of several domain generalization methods for both CNN and Transformer architectures, notably enhancing the state of the art. 
Newly demonstrating the power of LDMs as data generators for domain-robust segmentation, DGInStyle is one more step towards domain-independent semantic segmentation. We hope that it can lay the foundation for future work on how to best utilize generative models for improving domain generalization of dense scene understanding.

%
\bibliographystyle{splncs04}
\bibliography{main}
\clearpage
\setcounter{page}{1}
\setcounter{linenumber}{1}
\renewcommand*{\thesection}{\Alph{section}}
\newcommand{\multiref}[2]{\cref{#1}--\ref{#2}}
\renewcommand{\thetable}{S\arabic{table}}
\renewcommand{\thefigure}{S\arabic{figure}}
\setcounter{section}{0}
\setcounter{figure}{0}
\setcounter{table}{0}

In this supplementary document, we first present additional information about the diversity of the generated dataset in~\cref{supp:diversity}. We then provide a scale analysis of the dataset in~\cref{supp:scale}. In~\cref{supp:crcg},  detailed class-wise results of the proposed RCG are provided. The limitations of our approach are discussed in~\cref{supp:limitations}. Further example predictions are showcased in~\cref{supp:example-pred}, followed by additional examples of the MRLF module in~\cref{supp:example-mrlf} and samples in adverse weather conditions in~\cref{supp:adverse}.

\section{Diversity of the Generated Dataset}
\label{supp:diversity}
Our DGInStyle approach leverages the Style Swap and Style Prompting techniques to diversify the generated images. 
The diversity of training data is critical for the trained segmentation model's domain generalization. 
To further evaluate the diversity of the generated dataset, we employ the Frechet Inception Distance (FID)~\cite{heusel2017gans} and Kernel Inception Distance (KID)~\cite{binkowski2018demystifying}, which measure the distributional distance between two datasets. 
Specifically, we ablate the Style Swap and Style Prompting modules by assessing the similarity between our generated and five real-world datasets. 
The FID and KID scores are computed with~\cite{obukhov2020torchfidelity} and presented in~\cref{tab:style-fid} and~\cref{tab:style-kid}, respectively. 
A lower score indicates a smaller domain gap between the considered pair of datasets. 
Thus, a lower average score suggests a better coverage of the union of diverse datasets and, thus, better diversity of the generated data.
The results demonstrate that both components enhance the diversity of the generated data, with the highest quality attained when both are enabled. 

\begin{table}[]
\centering
\vspace{-5mm}
\renewcommand{\arraystretch}{1.1}
\setlength{\tabcolsep}{10pt}
\caption{
\textbf{
Quantitative evaluation of the generated data diversity
}%
using Frechet Inception Distance ($\downarrow$) between the generated data and real-world datasets.
Evidently, both Style Swap and Style Prompting play important roles in bridging the gap between the generated data and each of the real datasets, a union of which represents the task-specific domain of autonomous driving.
}
\vspace{-2mm}
\resizebox{1.0\columnwidth}{!}{
\begin{tabular}{cccccccc}
\toprule
Swap & Prompting & CS & BDD & MV & ACDC & DZ & Average\\
\midrule
\ding{56}  &  \ding{56} & 124.28& 98.57& 81.31 & 141.07 & 238.18 & 136.68     \\
\ding{52}  &  \ding{56} & 121.07 & 88.64 & 79.57 & 133.53 & 235.76 &129.71    \\
\ding{56}  &  \ding{52} & 121.98 & 95.25 & 80.02 & 136.21 & 233.97 & 133.48    \\
\ding{52}  &  \ding{52} & \textbf{117.05} & \textbf{88.46} & \textbf{74.81} &\textbf{128.39} & \textbf{227.69} & \textbf{127.37}  \\
\bottomrule        
\end{tabular}

}

\label{tab:style-fid}
\end{table}
\begin{table}[]
\centering
\vspace{-13mm}
\renewcommand{\arraystretch}{1.3}
\setlength{\tabcolsep}{5pt}
\caption{
\textbf{
Quantitative evaluation of the generated data diversity
}%
using Kernel Inception Distance (KID $\times$ 0.01 $\downarrow$) between the generated data and real-world datasets.
The standard deviation is part of the metric computation protocol and has also been scaled down by a factor of 0.01.
}
\vspace{-2mm}
\resizebox{1.0\textwidth}{!}{
\begin{tabular}{cccccccc}
\toprule
Swap & Prompting & CS & BDD & MV & ACDC & DZ & Average \\
\midrule

\ding{56}  &  \ding{56} & 8.54~$\pm$~0.15 & 5.62~$\pm$~0.08 & 4.99~$\pm$~0.14  & 7.95~$\pm$~0.18 & 15.66~$\pm$~0.54  & 8.55~$\pm$~0.22      \\
\ding{52}  &  \ding{56} & 8.19~$\pm$~0.19  & 4.98~$\pm$~0.09  & 5.00~$\pm$~0.15  & 7.40~$\pm$~0.16  & 15.38~$\pm$~0.53  & 8.19~$\pm$~0.23    \\
\ding{56}  &  \ding{52} & 8.24~$\pm$~0.20  & 5.41~$\pm$~0.08 & 5.04~$\pm$~0.13  & 7.50~$\pm$~0.18 & 14.93~$\pm$~0.64  & 8.23~$\pm$~0.24     \\
\ding{52}  &  \ding{52} & \textbf{7.86}~$\pm$~0.22  & \textbf{4.90}~$\pm$~0.09  & \textbf{4.98}~$\pm$~0.17  &\textbf{7.16}~$\pm$~0.18  & \textbf{14.36}~$\pm$~0.67  & \textbf{7.85}~$\pm$~0.27  \\
\bottomrule        
\end{tabular}
}

\label{tab:style-kid}
\end{table}

\section{Dataset scale analysis}
\label{supp:scale}
\cref{tab:supp-scale} studies the DG performance of DAFormer relative to the number of synthetic images. More generated images improve the mIoU up to around 6000 images, after which it reaches a plateau.
\begin{table}[]
\centering
\caption{
Performance of DAFormer Using DGInStyle wrt. the unmber of generated images (mIoU~$\uparrow$ in \%).
}
\footnotesize
\centering
\setlength{\tabcolsep}{9pt}
\begin{tabular}{@{}c cccccc @{}}
\toprule
$N_\mathcal{G}$  &   0   & 1000  & 2000 & 4000  & 6000  & 8000  \\
\hline
Avg3 & 51.73 & 53.57 &  53.86 &  54.1  & 54.25 &  54.28 \\
Avg5 & 42.18 & 44.95 & 45.86   & 46.22 & 46.47 &  46.39 \\
\bottomrule
\end{tabular}
\label{tab:supp-scale}
\end{table}

\section{Class-wise results of RCG}
\label{supp:crcg}
\begin{figure}[]
    \centering

    \includegraphics[width=\linewidth,trim={0em 0.5em 0em 0em},clip]{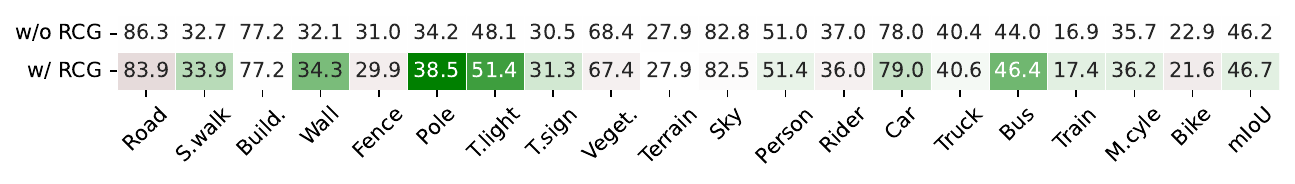}    
    \caption{
    Comparison of the class-wise IoU averaged over the
        five datasets with and without RCG while keeping the other 
        components of DGInStyle coupled with DAFormer. The color 
        visualizes the difference to the first row.}    
    \label{fig:rcg_ciou_heatmap}

\end{figure}

In~\cref{fig:rcg_ciou_heatmap}, we show the effectiveness of RCG for difficult classes, such as \textit{pole}, \textit{traffic light} and \textit{bus} that have a low pixel count in the source data.

\section{Limitations}
\label{supp:limitations}
Diffusion models exhibit a primary drawback of prolonged sampling times. As our model is based on diffusion models, it naturally inherits this slow inference property. Moreover, the proposed MRLF module operates on multiple tiles cropped from the upscaled latents, and the sampling process of all these tiles further extends the image generation duration. However, it is important to note that this extended diffusion time does not impact the inference time of the deployed segmentation networks. Furthermore, much ongoing research aims to expedite diffusion model sampling, and we believe that this issue can be alleviated through architectural advancements.

\section{Further Example Predictions}
\label{supp:example-pred}
\begin{figure}[]
    \centering
    \includegraphics[width=0.8\linewidth,trim={0 0 0 0},clip]{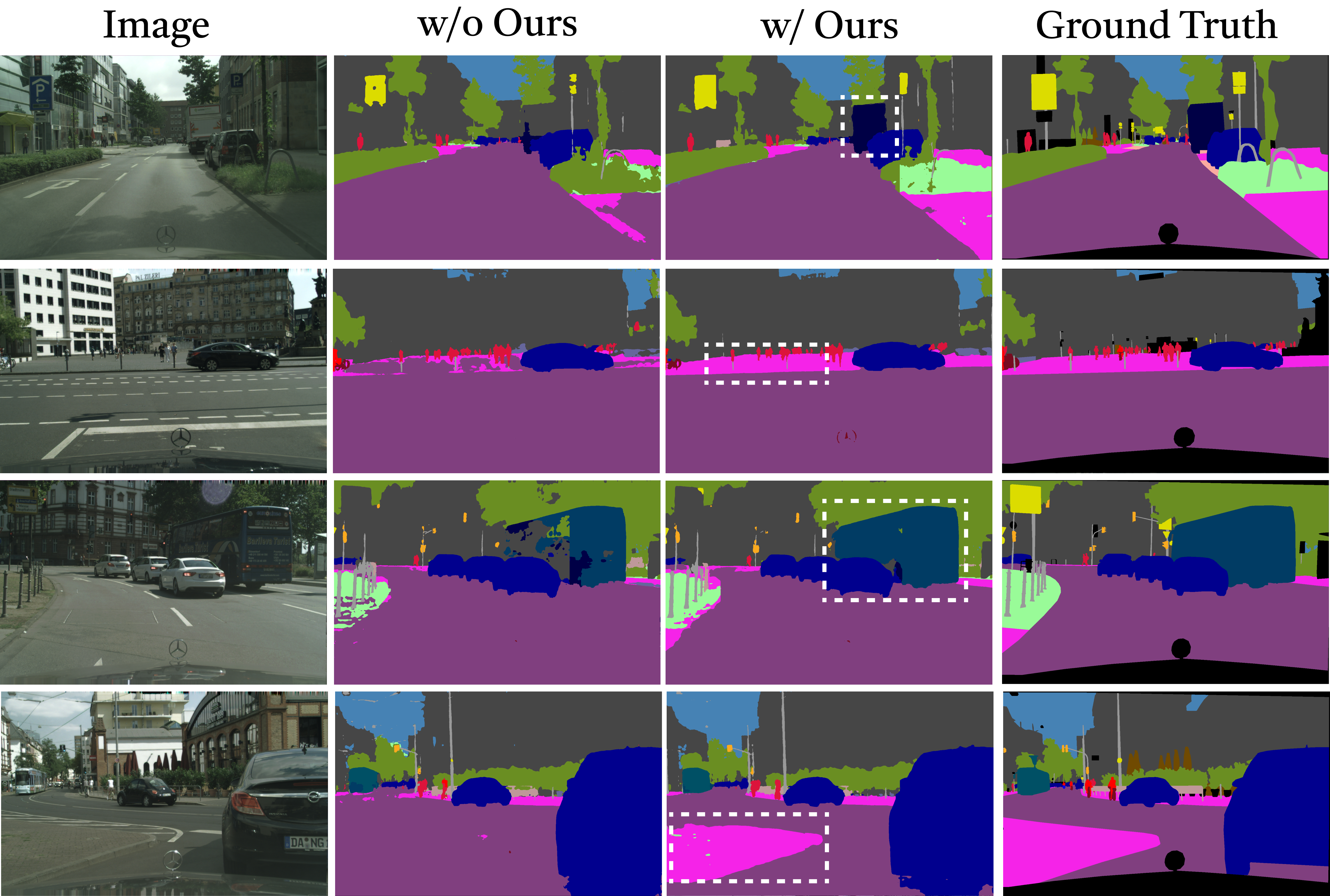}
     \resizebox{1.0\linewidth}{1.4mm}{\scriptsize%
\setlength\tabcolsep{1pt}%
{%
\newcolumntype{P}[1]{>{\centering\arraybackslash}p{#1}}
\begin{tabular}{@{}*{20}{P{0.09\columnwidth}}@{}}
     {\cellcolor[rgb]{0.5,0.25,0.5}}\textcolor{white}{road} 
     &{\cellcolor[rgb]{0.957,0.137,0.91}}sidew. 
     &{\cellcolor[rgb]{0.275,0.275,0.275}}\textcolor{white}{build.} 
     &{\cellcolor[rgb]{0.4,0.4,0.612}}\textcolor{white}{wall} 
     &{\cellcolor[rgb]{0.745,0.6,0.6}}fence 
     &{\cellcolor[rgb]{0.6,0.6,0.6}}pole 
     &{\cellcolor[rgb]{0.98,0.667,0.118}}tr. light
     &{\cellcolor[rgb]{0.863,0.863,0}}tr. sign
     &{\cellcolor[rgb]{0.42,0.557,0.137}}veget. 
     &{\cellcolor[rgb]{0.596,0.984,0.596}}terrain 
     &{\cellcolor[rgb]{0.275,0.510,0.706}}sky
     &{\cellcolor[rgb]{0.863,0.078,0.235}}\textcolor{white}{person} 
     &{\cellcolor[rgb]{0.988,0.494,0.635}}\textcolor{black}{rider} 
     &{\cellcolor[rgb]{0,0,0.557}}\textcolor{white}{car} 
     &{\cellcolor[rgb]{0,0,0.275}}\textcolor{white}{truck} 
     &{\cellcolor[rgb]{0,0.235,0.392}}\textcolor{white}{bus}
     &{\cellcolor[rgb]{0,0.392,0.471}}\textcolor{white}{train} 
     &{\cellcolor[rgb]{0,0,0.902}}\textcolor{white}{m.bike} 
     & {\cellcolor[rgb]{0.467,0.043,0.125}}\textcolor{white}{bike}
     &{\cellcolor[rgb]{0,0,0}}\textcolor{white}{n/a.}
\end{tabular}
}%
}
    \caption{Example predictions from HRDA trained with and w/o our DGInStyle on the \textbf{Cityscapes} dataset, showing improved performance on \textit{truck} and \textit{bus} and exhibiting a more complete segmentation of \textit{sidewalk}.}
    \label{fig:pred-examples-cs}
\end{figure}

\begin{figure}[]
    \centering
    \includegraphics[width=0.8\linewidth,trim={0 0 0 0},clip]{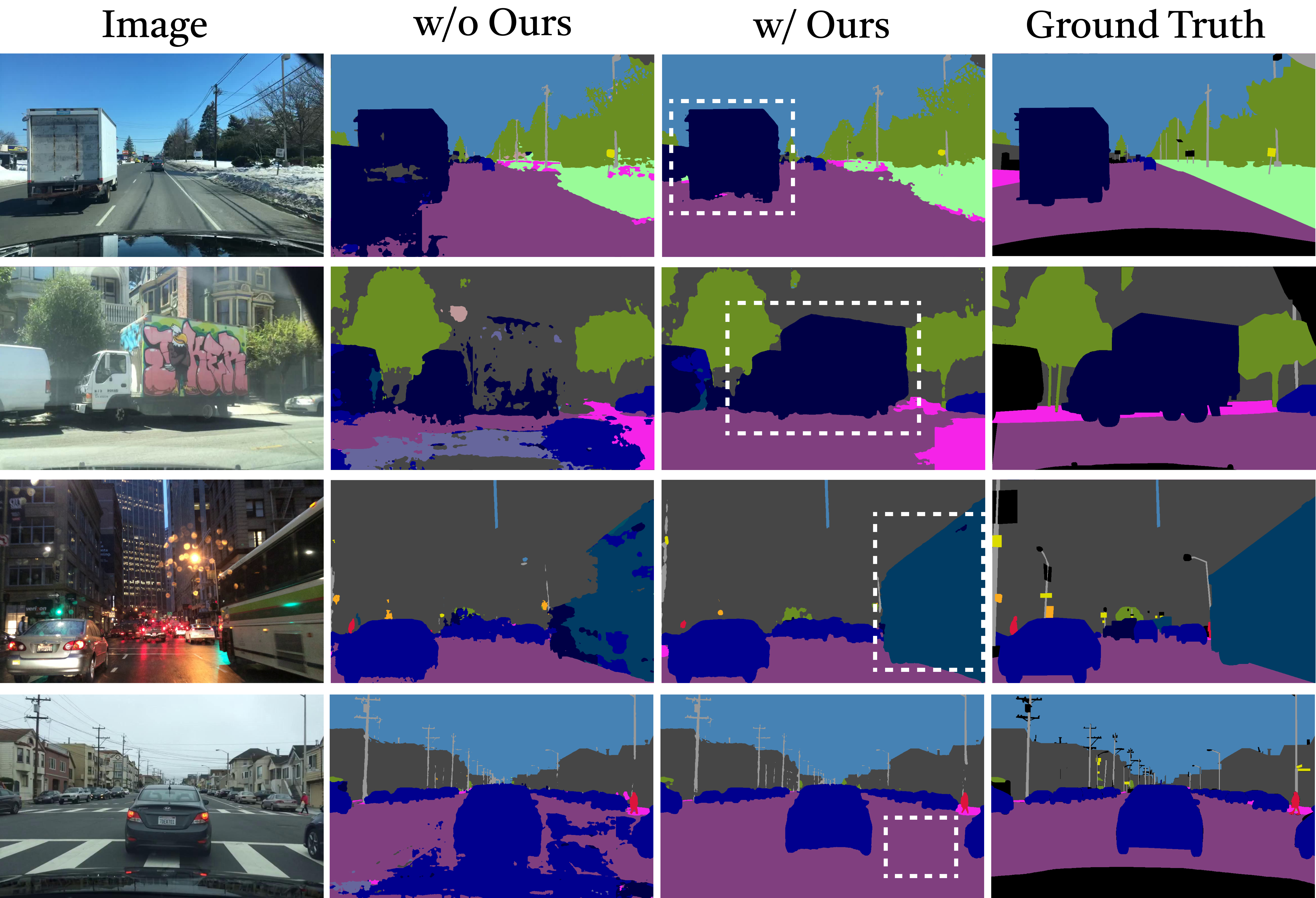}
     \resizebox{1.0\linewidth}{1.4mm}{\scriptsize%
\setlength\tabcolsep{1pt}%
{%
\newcolumntype{P}[1]{>{\centering\arraybackslash}p{#1}}
\begin{tabular}{@{}*{20}{P{0.09\columnwidth}}@{}}
     {\cellcolor[rgb]{0.5,0.25,0.5}}\textcolor{white}{road} 
     &{\cellcolor[rgb]{0.957,0.137,0.91}}sidew. 
     &{\cellcolor[rgb]{0.275,0.275,0.275}}\textcolor{white}{build.} 
     &{\cellcolor[rgb]{0.4,0.4,0.612}}\textcolor{white}{wall} 
     &{\cellcolor[rgb]{0.745,0.6,0.6}}fence 
     &{\cellcolor[rgb]{0.6,0.6,0.6}}pole 
     &{\cellcolor[rgb]{0.98,0.667,0.118}}tr. light
     &{\cellcolor[rgb]{0.863,0.863,0}}tr. sign
     &{\cellcolor[rgb]{0.42,0.557,0.137}}veget. 
     &{\cellcolor[rgb]{0.596,0.984,0.596}}terrain 
     &{\cellcolor[rgb]{0.275,0.510,0.706}}sky
     &{\cellcolor[rgb]{0.863,0.078,0.235}}\textcolor{white}{person} 
     &{\cellcolor[rgb]{0.988,0.494,0.635}}\textcolor{black}{rider} 
     &{\cellcolor[rgb]{0,0,0.557}}\textcolor{white}{car} 
     &{\cellcolor[rgb]{0,0,0.275}}\textcolor{white}{truck} 
     &{\cellcolor[rgb]{0,0.235,0.392}}\textcolor{white}{bus}
     &{\cellcolor[rgb]{0,0.392,0.471}}\textcolor{white}{train} 
     &{\cellcolor[rgb]{0,0,0.902}}\textcolor{white}{m.bike} 
     & {\cellcolor[rgb]{0.467,0.043,0.125}}\textcolor{white}{bike}
     &{\cellcolor[rgb]{0,0,0}}\textcolor{white}{n/a.}
\end{tabular}
}%
}
    \caption{Example predictions from HRDA trained with and w/o our DGInStyle on the \textbf{BDD100K} dataset, showing a better recognition of difficult classes such as \textit{truck} and \textit{bus}.}
    \label{fig:pred-examples-bdd}
\end{figure}

\begin{figure}[]
    \centering
    \includegraphics[width=0.8\linewidth,trim={0 0 0 0},clip]{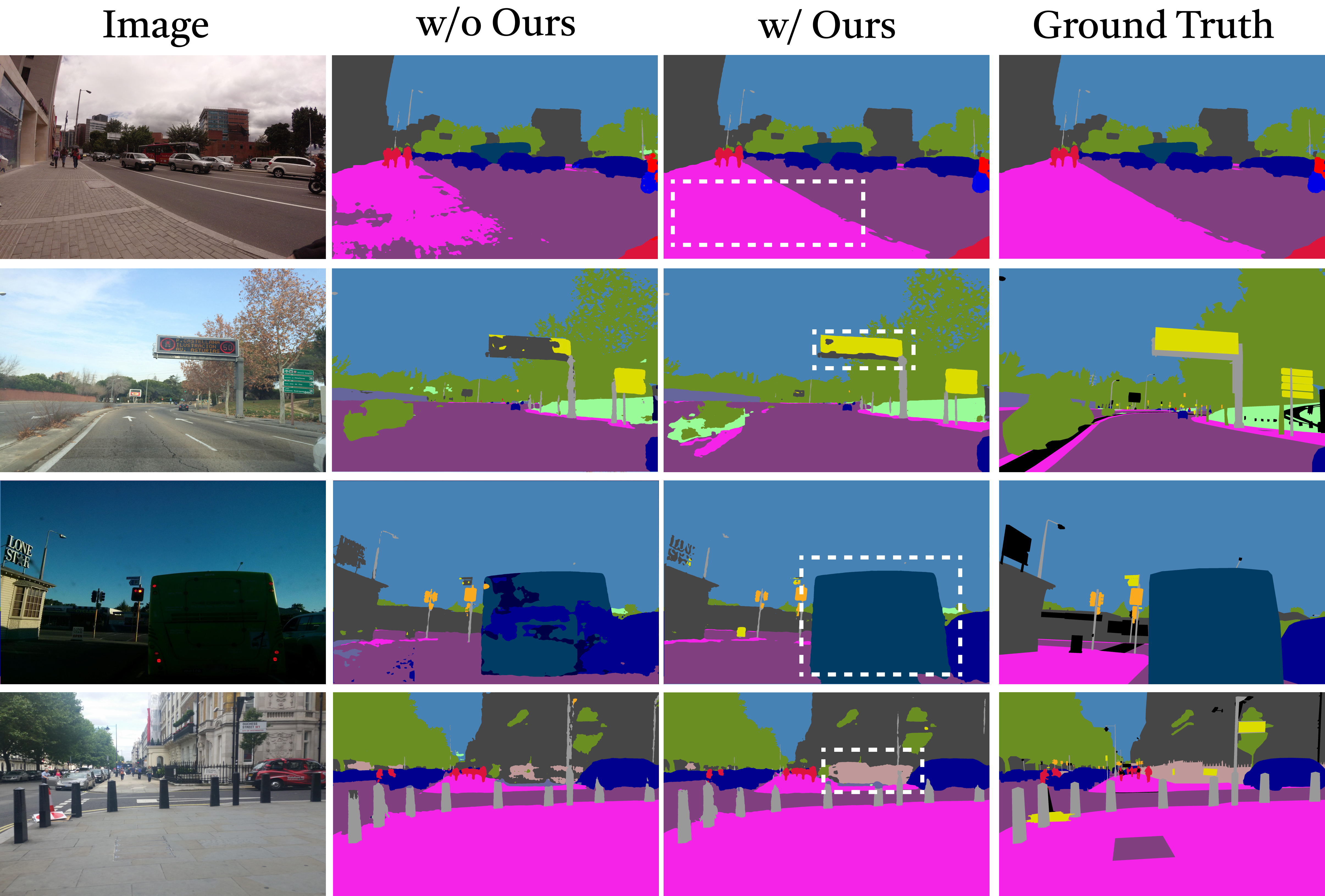}
     \resizebox{1.0\linewidth}{1.4mm}{\scriptsize%
\setlength\tabcolsep{1pt}%
{%
\newcolumntype{P}[1]{>{\centering\arraybackslash}p{#1}}
\begin{tabular}{@{}*{20}{P{0.09\columnwidth}}@{}}
     {\cellcolor[rgb]{0.5,0.25,0.5}}\textcolor{white}{road} 
     &{\cellcolor[rgb]{0.957,0.137,0.91}}sidew. 
     &{\cellcolor[rgb]{0.275,0.275,0.275}}\textcolor{white}{build.} 
     &{\cellcolor[rgb]{0.4,0.4,0.612}}\textcolor{white}{wall} 
     &{\cellcolor[rgb]{0.745,0.6,0.6}}fence 
     &{\cellcolor[rgb]{0.6,0.6,0.6}}pole 
     &{\cellcolor[rgb]{0.98,0.667,0.118}}tr. light
     &{\cellcolor[rgb]{0.863,0.863,0}}tr. sign
     &{\cellcolor[rgb]{0.42,0.557,0.137}}veget. 
     &{\cellcolor[rgb]{0.596,0.984,0.596}}terrain 
     &{\cellcolor[rgb]{0.275,0.510,0.706}}sky
     &{\cellcolor[rgb]{0.863,0.078,0.235}}\textcolor{white}{person} 
     &{\cellcolor[rgb]{0.988,0.494,0.635}}\textcolor{black}{rider} 
     &{\cellcolor[rgb]{0,0,0.557}}\textcolor{white}{car} 
     &{\cellcolor[rgb]{0,0,0.275}}\textcolor{white}{truck} 
     &{\cellcolor[rgb]{0,0.235,0.392}}\textcolor{white}{bus}
     &{\cellcolor[rgb]{0,0.392,0.471}}\textcolor{white}{train} 
     &{\cellcolor[rgb]{0,0,0.902}}\textcolor{white}{m.bike} 
     & {\cellcolor[rgb]{0.467,0.043,0.125}}\textcolor{white}{bike}
     &{\cellcolor[rgb]{0,0,0}}\textcolor{white}{n/a.}
\end{tabular}
}%
}
    \caption{Example predictions from HRDA trained with and w/o our DGInStyle on the \textbf{Mapillary Vistas} dataset, showing an improved performance of \textit{sidewalk}, \textit{traffic sign}, \textit{bus} and \textit{fence}.}
    \label{fig:pred-examples-mv}
\end{figure}

\begin{figure}[]
    \centering
    \includegraphics[width=0.8\linewidth,trim={0 0 0 0},clip]{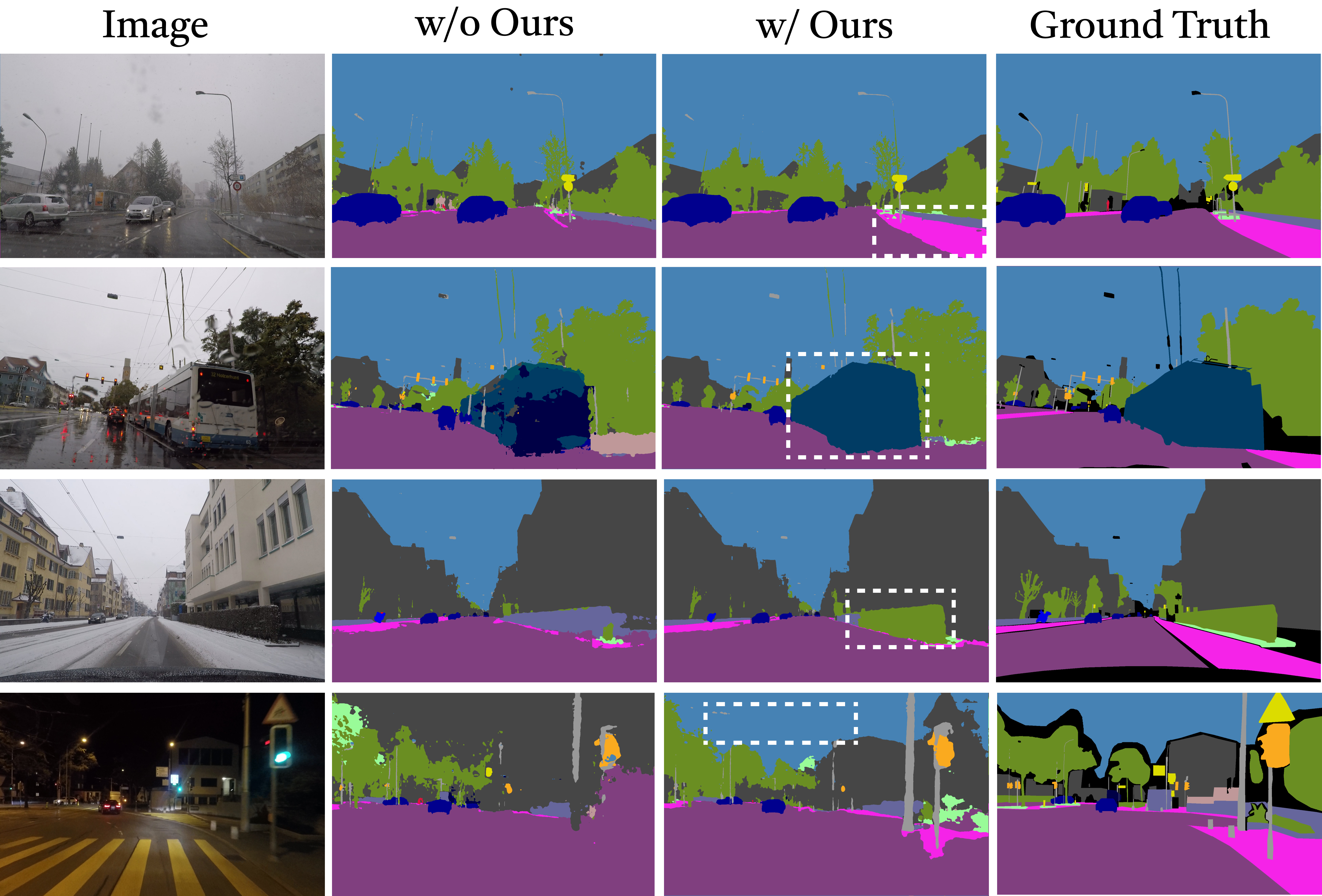}
    
    \resizebox{1.0\linewidth}{1.4mm}{\scriptsize%
\setlength\tabcolsep{1pt}%
{%
\newcolumntype{P}[1]{>{\centering\arraybackslash}p{#1}}
\begin{tabular}{@{}*{20}{P{0.09\columnwidth}}@{}}
     {\cellcolor[rgb]{0.5,0.25,0.5}}\textcolor{white}{road} 
     &{\cellcolor[rgb]{0.957,0.137,0.91}}sidew. 
     &{\cellcolor[rgb]{0.275,0.275,0.275}}\textcolor{white}{build.} 
     &{\cellcolor[rgb]{0.4,0.4,0.612}}\textcolor{white}{wall} 
     &{\cellcolor[rgb]{0.745,0.6,0.6}}fence 
     &{\cellcolor[rgb]{0.6,0.6,0.6}}pole 
     &{\cellcolor[rgb]{0.98,0.667,0.118}}tr. light
     &{\cellcolor[rgb]{0.863,0.863,0}}tr. sign
     &{\cellcolor[rgb]{0.42,0.557,0.137}}veget. 
     &{\cellcolor[rgb]{0.596,0.984,0.596}}terrain 
     &{\cellcolor[rgb]{0.275,0.510,0.706}}sky
     &{\cellcolor[rgb]{0.863,0.078,0.235}}\textcolor{white}{person} 
     &{\cellcolor[rgb]{0.988,0.494,0.635}}\textcolor{black}{rider} 
     &{\cellcolor[rgb]{0,0,0.557}}\textcolor{white}{car} 
     &{\cellcolor[rgb]{0,0,0.275}}\textcolor{white}{truck} 
     &{\cellcolor[rgb]{0,0.235,0.392}}\textcolor{white}{bus}
     &{\cellcolor[rgb]{0,0.392,0.471}}\textcolor{white}{train} 
     &{\cellcolor[rgb]{0,0,0.902}}\textcolor{white}{m.bike} 
     & {\cellcolor[rgb]{0.467,0.043,0.125}}\textcolor{white}{bike}
     &{\cellcolor[rgb]{0,0,0}}\textcolor{white}{n/a.}
\end{tabular}
}%
}
    \caption{Example predictions from HRDA trained with and w/o our DGInStyle  on the \textbf{ACDC} dataset, demonstrating improved performance in rainy and snowy conditions for classes such as \textit{sidewalk}, \textit{bus}, \textit{vegetation} and \textit{sky}.}
    \label{fig:pred-examples-acdc}
\end{figure}

\begin{figure}[]
    \centering
    \includegraphics[width=0.8\linewidth,trim={0 0 0 0},clip]{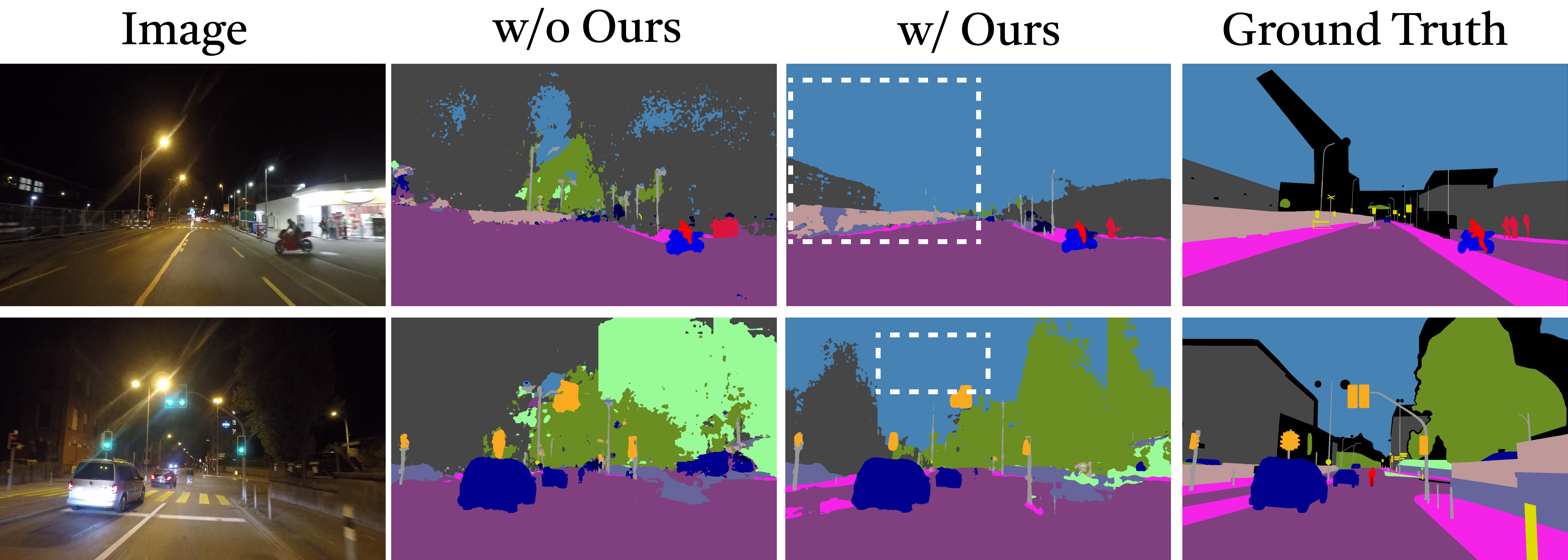}
     \resizebox{1.0\linewidth}{1.4mm}{\scriptsize%
\setlength\tabcolsep{1pt}%
{%
\newcolumntype{P}[1]{>{\centering\arraybackslash}p{#1}}
\begin{tabular}{@{}*{20}{P{0.09\columnwidth}}@{}}
     {\cellcolor[rgb]{0.5,0.25,0.5}}\textcolor{white}{road} 
     &{\cellcolor[rgb]{0.957,0.137,0.91}}sidew. 
     &{\cellcolor[rgb]{0.275,0.275,0.275}}\textcolor{white}{build.} 
     &{\cellcolor[rgb]{0.4,0.4,0.612}}\textcolor{white}{wall} 
     &{\cellcolor[rgb]{0.745,0.6,0.6}}fence 
     &{\cellcolor[rgb]{0.6,0.6,0.6}}pole 
     &{\cellcolor[rgb]{0.98,0.667,0.118}}tr. light
     &{\cellcolor[rgb]{0.863,0.863,0}}tr. sign
     &{\cellcolor[rgb]{0.42,0.557,0.137}}veget. 
     &{\cellcolor[rgb]{0.596,0.984,0.596}}terrain 
     &{\cellcolor[rgb]{0.275,0.510,0.706}}sky
     &{\cellcolor[rgb]{0.863,0.078,0.235}}\textcolor{white}{person} 
     &{\cellcolor[rgb]{0.988,0.494,0.635}}\textcolor{black}{rider} 
     &{\cellcolor[rgb]{0,0,0.557}}\textcolor{white}{car} 
     &{\cellcolor[rgb]{0,0,0.275}}\textcolor{white}{truck} 
     &{\cellcolor[rgb]{0,0.235,0.392}}\textcolor{white}{bus}
     &{\cellcolor[rgb]{0,0.392,0.471}}\textcolor{white}{train} 
     &{\cellcolor[rgb]{0,0,0.902}}\textcolor{white}{m.bike} 
     & {\cellcolor[rgb]{0.467,0.043,0.125}}\textcolor{white}{bike}
     &{\cellcolor[rgb]{0,0,0}}\textcolor{white}{n/a.}
\end{tabular}
}%
}
    \caption{Example predictions from HRDA trained with and w/o our DGInStyle on the \textbf{Dark Zurich} dataset, demonstrating superior generalization for dark scenes in the \textit{sky} and \textit{vegetation} classes.}
    \label{fig:pred-examples-dz}
\end{figure}

We present a comprehensive qualitative comparison between the predicted semantic segmentation results of HRDA trained with GTA-only data and the model trained with our DGInStyle approach. 
We evaluate these models on real-world datasets, including Cityscapes (\cf~\cref{fig:pred-examples-cs}), BDD100K (\cf~\cref{fig:pred-examples-bdd}), Mapillary Vistas (\cf~\cref{fig:pred-examples-mv}), ACDC (\cf~\cref{fig:pred-examples-acdc}), and Dark Zurich (\cf~\cref{fig:pred-examples-dz}). 
The model trained with our DGInStyle can better segment \textit{truck} and \textit{bus} (as seen in~\multiref{fig:pred-examples-cs}{fig:pred-examples-acdc}). 
It also exhibits a correct segmentation of \textit{sidewalk}, effectively identifying areas that were previously overlooked by the GTA-only trained model (as seen in~\cref{fig:pred-examples-cs}, \cref{fig:pred-examples-mv}). 
Furthermore, it enhances performance for rare classes, such as \textit{fence} and \textit{traffic sign} (as seen in~\cref{fig:pred-examples-mv}). 
In challenging conditions, such as nighttime scenes, our DGInStyle approach significantly improves the segmentation of \textit{sky} and \textit{vegetation} (as seen in~\cref{fig:pred-examples-acdc} and~\cref{fig:pred-examples-dz}).
\vspace{10mm}


\afterpage{
\section{Multi-Resolution Latent Fusion Module}
\label{supp:example-mrlf}

In \multiref{fig:mrlf1}{fig:mrlf3}, we provide additional qualitative examples showing how the MRLF module mitigates issues of the base Stable Diffusion LDM related to the poor quality of small objects generation. For instance, in~\cref{fig:mrlf1} (a), the motorcycle and rider are initially indistinct and poorly rendered. However, after applying the MRLF module, these elements become clearly recognizable and well-defined. Similarly, the fine-grained poles' details show a marked improvement in~\cref{fig:mrlf2}. Additionally, the quality of the person depicted in~\cref{fig:mrlf3} also benefits significantly from the MRLF module, demonstrating its overall effectiveness in refining and improving the quality of small-scale features in generated images.

\begin{figure}[h]
    \centering
    \includegraphics[width=0.8\linewidth,trim={0 0 0 0},clip]{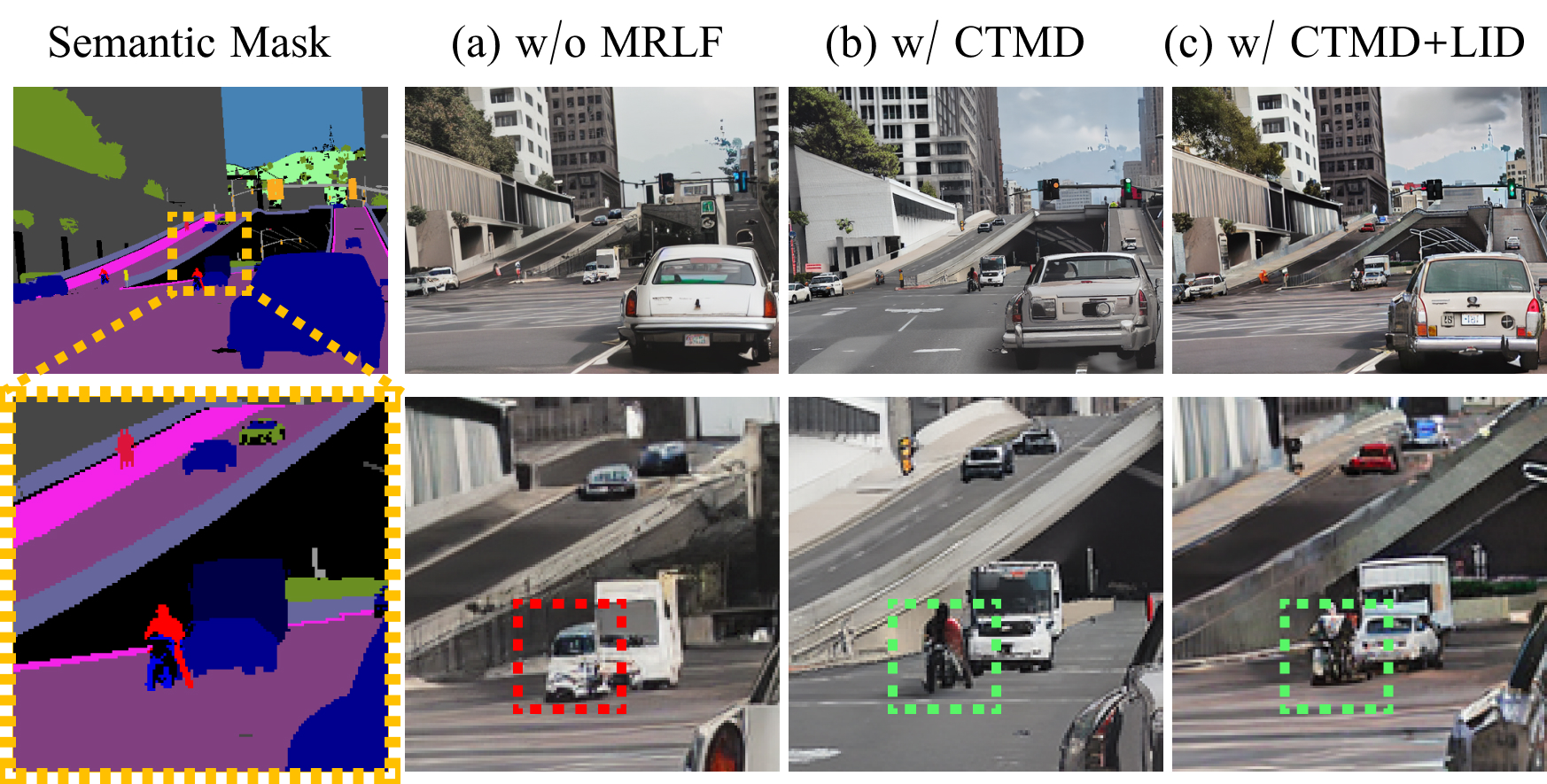}
    \caption{Qualitative example of MRLF: improved generation for small distant objects like \textit{rider} and \textit{motorcycle} when zooming in on the mask crop.}
    \label{fig:mrlf1}
\end{figure}
\begin{figure}[h]
    \centering
    \includegraphics[width=0.8\linewidth,trim={0 0 0 0},clip]{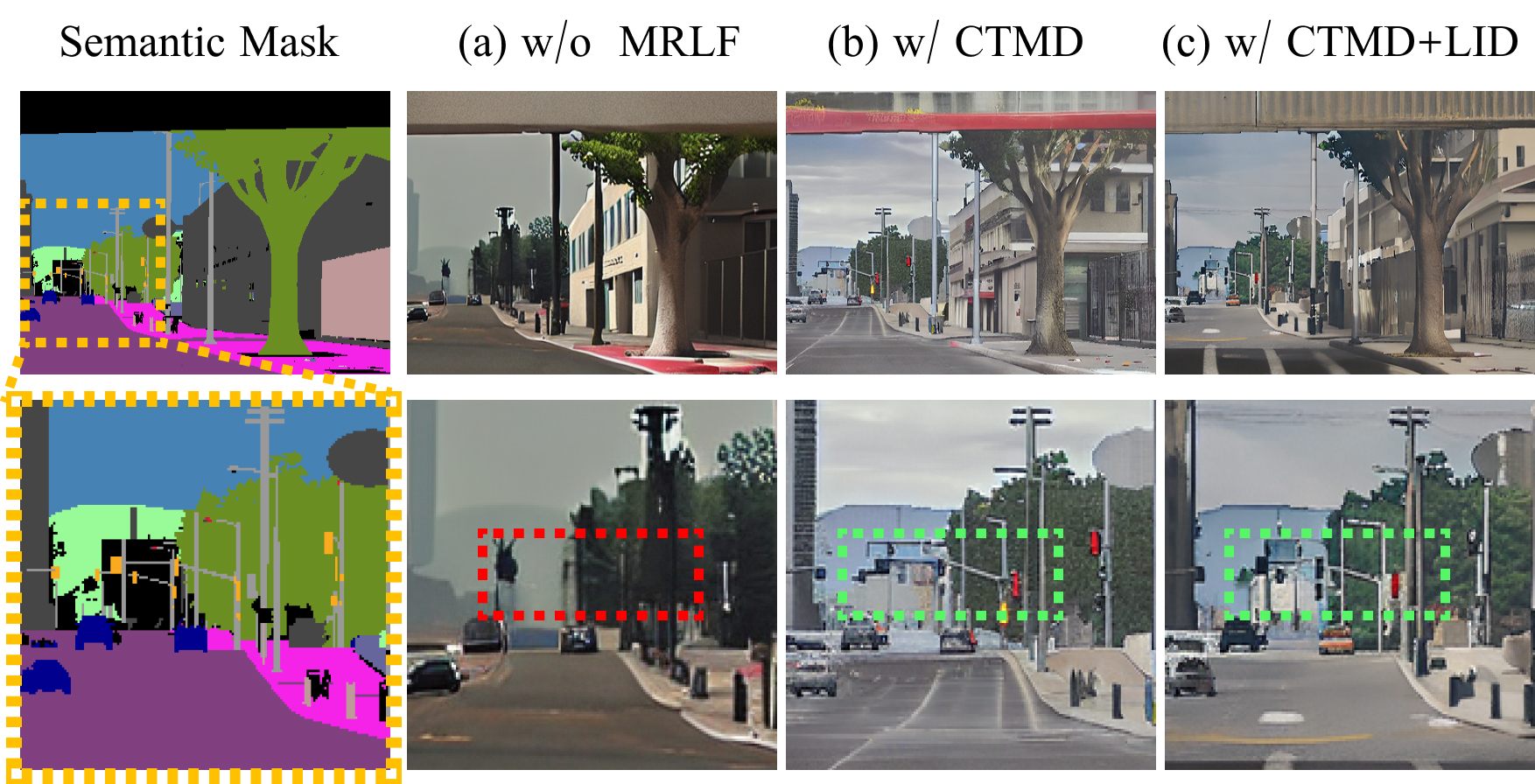}
    \caption{Qualitative example of MRLF: improved generation for small distant objects like \textit{pole} and \textit{traffic light} when zooming in on the mask crop.}
    \label{fig:mrlf2}
\end{figure}
\begin{figure}[h]
    \centering
    \includegraphics[width=0.8\linewidth,trim={0 0 0 0},clip]{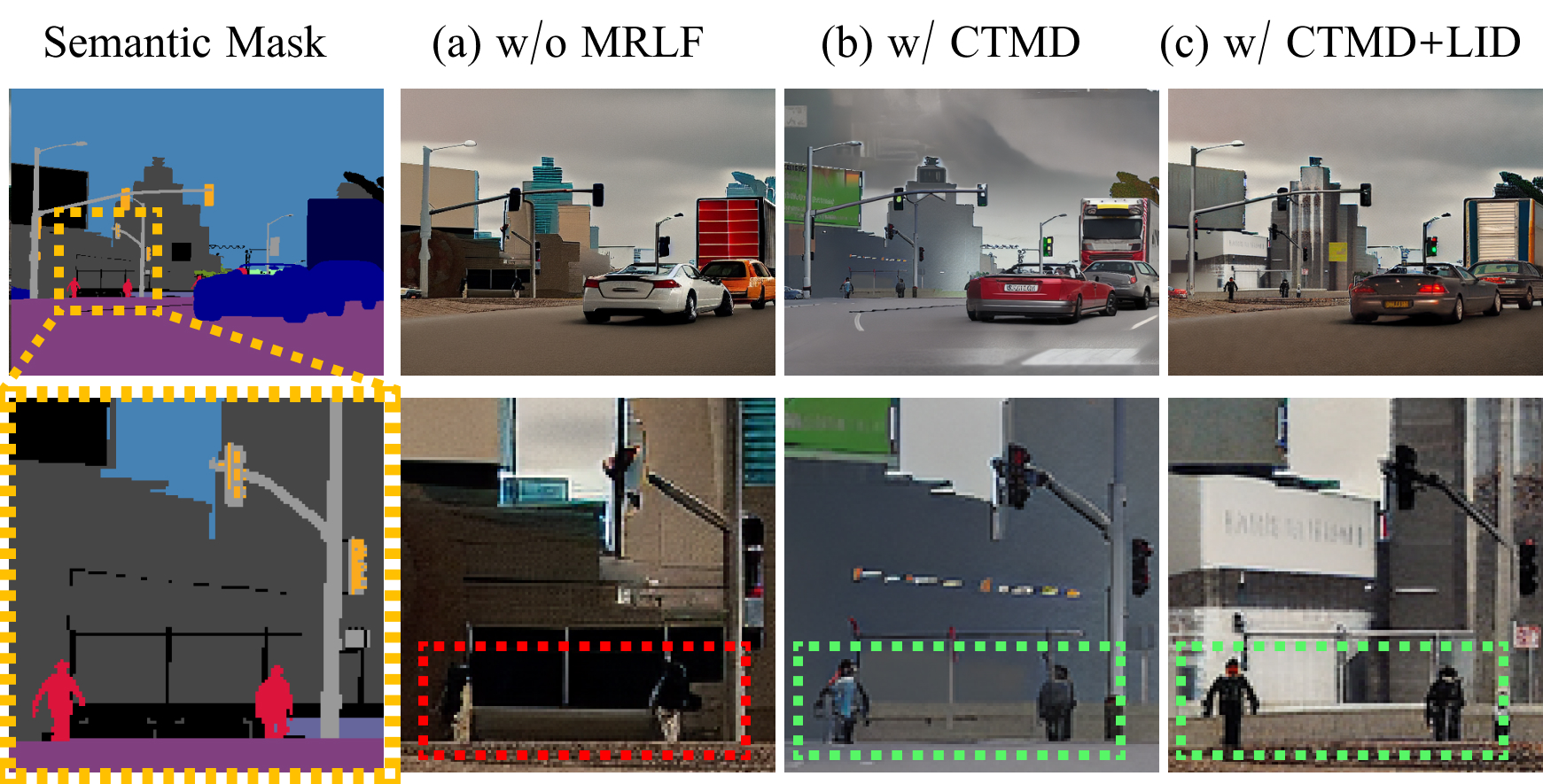}
    \caption{Qualitative example of MRLF: improved generation for small distant objects like \textit{person} when zooming in on the mask crop.}
    \label{fig:mrlf3}
    \vspace{-8mm}
\end{figure}
}

\afterpage{
\section{Adverse Weather Samples }
\label{supp:adverse}
In~\cref{fig:weather-samples}, we show more examples of the generated content under different weather conditions given the same semantic label condition. 
By encompassing a wide range of weather scenarios, DGInStyle ensures that the models are well-equipped to handle real-world variations, thereby improving their applicability and reliability in diverse operational environments.
\begin{figure*}[b!]
    \vspace{-10mm}
    \centering
    \includegraphics[width=1.0\linewidth,trim={0 0 0 0},clip]{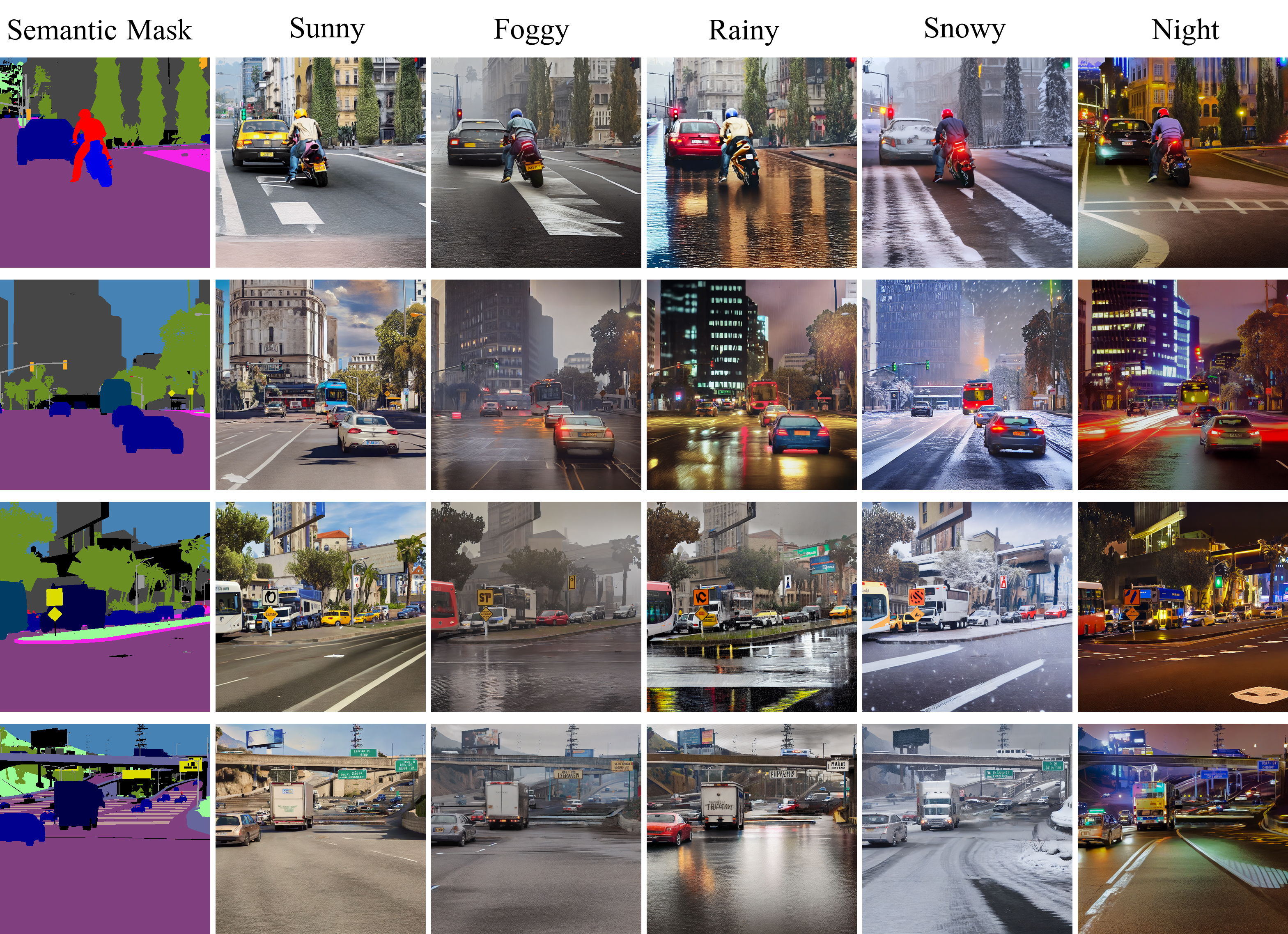}
    \caption{Examples generated by our DGInStyle approach under varying weather conditions, all based on the same semantic label condition. }
    \label{fig:weather-samples}
    \vspace{-10mm}
\end{figure*}
}

\end{document}